\documentclass[runningheads]{fmcad}

\usepackage{cite}
\usepackage{amsmath,amssymb,amsfonts}
\usepackage[ruled,linesnumbered,noend]{algorithm2e}

\usepackage{graphicx}
\usepackage{subcaption}
\usepackage{textcomp}

\usepackage{alltt}
\usepackage{amssymb}
\usepackage{caption}
\usepackage{paralist}
\usepackage{nicefrac}
\usepackage{booktabs}
\usepackage{xspace}
\usepackage{amsmath}
\usepackage{marvosym}
\usepackage{wasysym}
\usepackage{verbatim}
\usepackage{float}
\usepackage{hyperref}
\usepackage{multirow}
\usepackage{cite}
\usepackage[capitalize]{cleveref}
\usepackage[per-mode=symbol,detect-all]{siunitx}
\usepackage{graphics}
\usepackage{amssymb}
\usepackage{enumitem}

\usepackage{booktabs}
\usepackage{colortbl}
\usepackage{ifthen}
\usepackage{mathdots}
\usepackage{arydshln}
\usepackage[outline]{contour}

\usepackage{wrapfig}
\usepackage{MnSymbol}
\usepackage{placeins}
\usepackage{diagbox}
\usepackage{listings}
\usepackage{pdfpages}

\usepackage{xcolor,pifont}
\newcommand*\colourcheck[1]{%
	\expandafter\newcommand\csname #1check\endcsname{\textcolor{#1}{\ding{52}}}%
}
\newcommand*\colourxmark[1]{%
	\expandafter\newcommand\csname #1xmark\endcsname{\textcolor{#1}{\ding{56}}}%
}
\colourcheck{green}
\colourxmark{red}

\definecolor{navy}{RGB}{0,0,128}

\usepackage{tikz,ifthen,pgfplots}
\usetikzlibrary{arrows,trees,backgrounds,automata,shapes,decorations,plotmarks,fit,calc,positioning,shadows,chains}
\usetikzlibrary{quotes,arrows.meta}
\tikzstyle{every pin edge}=[<-,shorten <=1pt]
\tikzstyle{neuron}=[circle,fill=black!25,minimum size=17pt,inner sep=0pt]
\tikzstyle{input neuron}=[neuron, fill=green!50]
\tikzstyle{output neuron}=[neuron, fill=red!50]
\tikzstyle{hidden neuron}=[neuron, fill=blue!50]
\tikzstyle{small neuron}        =[hidden neuron, draw, minimum size=15pt]
\tikzstyle{small input neuron}  =[input neuron , draw, minimum size=15pt]
\tikzstyle{small output neuron} =[output neuron, draw, minimum size=15pt]
\tikzstyle{annot} = [text width=4em, text centered]
\tikzstyle{nnedge} = [-{stealth},shorten >=0.1cm, shorten <=0.05cm,line width=0.8pt,black]
\tikzstyle{edge} = [->,line width = 0.3pt, shorten >=0.2cm]
\tikzstyle{edgeWide} = [->,line width = 2pt, , shorten >=0.2cm]
\usetikzlibrary{calc}

\tikzset{every picture/.style={line width=0.75pt}} 
\tikzstyle{BadSquare}=[rectangle,fill=red!30!white,minimum size=25pt,inner 
sep=0pt]
\tikzstyle{InitSquare}=[rectangle,fill=green!30!white,minimum size=25pt,inner 
sep=0pt]

\newcommand{\sat}{\texttt{SAT}\xspace}

\newcommand{\unsat}{\texttt{UNSAT}\xspace}

\usepackage{amsthm}
\newtheorem{lemma}{Lemma}

\usepackage{mdframed}

\newmdtheoremenv{definition}{Definition}

\usepackage{bussproofs}
\usepackage{gensymb}

\newif\ifcomments
\commentstrue

\newif\ifoutline
\outlinefalse

\newif\iflong
\longtrue

\renewcommand{\paragraph}[1]{\vspace{1mm}\noindent{\bf #1}\ }

\newcommand{\states}{\mathcal{X}\xspace}
\newcommand{\state}{x\xspace}
\newcommand{\traj}{\tau\xspace}
\newcommand{\control}{u\xspace}
\newcommand{\controls}{\mathcal{U}\xspace}
\newcommand{\trans}{f\xspace}
\newcommand{\certfun}{V\xspace}
\newcommand{\ctrlfun}{\pi\xspace}

\begin{document}

\IEEEoverridecommandlockouts
\DeclareRobustCommand{\IEEEauthorrefmark}[1]{\smash{\textsuperscript{\footnotesize
 #1}}}

	
\title{Formally Verifying Deep Reinforcement Learning\\ Controllers with Lyapunov Barrier Certificates } 
	
	\author{
		\IEEEauthorblockN{
			Udayan Mandal\IEEEauthorrefmark{1},
			Guy Amir\IEEEauthorrefmark{2},
			 Haoze Wu\IEEEauthorrefmark{1},
			 Ieva Daukantas\IEEEauthorrefmark{3},
            Fletcher Lee Newell\IEEEauthorrefmark{1},
			Umberto J. Ravaioli\IEEEauthorrefmark{4},\\
			 Baoluo Meng\IEEEauthorrefmark{5},
			 Michael Durling\IEEEauthorrefmark{5},
          Milan Ganai\IEEEauthorrefmark{1},
          Tobey 
          Shim\IEEEauthorrefmark{1},
          Guy Katz\IEEEauthorrefmark{2},
			and Clark Barrett\IEEEauthorrefmark{1}
		}
  
		\IEEEauthorblockA{			\IEEEauthorrefmark{1}Stanford University, 
			\IEEEauthorrefmark{2}The Hebrew University of Jerusalem, 
   \IEEEauthorrefmark{3}IT University of Copenhagen,
\IEEEauthorrefmark{4}Google, 
   \IEEEauthorrefmark{5}GE Aerospace Research
		}
	}

\maketitle

\begin{abstract}
  Deep reinforcement learning (DRL) is a powerful machine learning paradigm for generating agents that control autonomous systems. 
  However, the ``black box'' nature of DRL agents
  limits their deployment in real-world safety-critical applications. 
  A promising approach for providing strong guarantees on an agent's behavior is to use \emph{Neural Lyapunov Barrier} (NLB) certificates, which are learned functions over the system  whose properties indirectly imply that an agent behaves as desired. 
  However, NLB-based certificates are typically difficult to learn and even more difficult to verify, especially for complex systems.
  In this work, we present a novel method for training and verifying NLB-based certificates for discrete-time systems. 
  Specifically, we introduce a technique for certificate \emph{composition}, which simplifies the verification of highly-complex systems by strategically designing a sequence of certificates.  When jointly verified with neural network verification engines, these certificates provide a formal guarantee that a DRL agent both achieves its goals and avoids unsafe behavior. 
  Furthermore, we introduce a technique for certificate \emph{filtering}, which significantly simplifies the process of producing formally verified certificates.
  We demonstrate the merits of our approach with a case study on providing safety and liveness guarantees for a DRL-controlled spacecraft.
  \end{abstract}

\section{Introduction}
\label{sec:Introduction}
In recent years, deep reinforcement learning (DRL) has achieved unprecedented results in multiple domains, including game playing, robotic control, protein folding, and many more \cite{cao2024survey, hester2011realtime, 10.1145/3569966.3570102, mnih2013playing}.
However, such models have an opaque decision-making process, making it highly challenging to determine whether a DRL-based system will \emph{always} behave correctly. 
This is especially concerning for safety-critical domains (e.g., autonomous vehicles), in which even a single mistake can have dire consequences and risk human lives. 
This drawback limits the incorporation of DRL in real-world safety-critical systems. 

The formal methods community has responded to this challenge by developing automated reasoning approaches for \emph{proving} that a DRL-based controller behaves correctly \cite{10.1145/3596444} . These efforts rely in part on specialized DNN verification engines (a.k.a. \emph{DNN verifiers}), which adapt techniques from other domains such as satisfiability modulo theories, abstract interpretation, mixed integer linear programming, and convex optimization \cite{KaBaDiJuKo17, KaBaDiJuKo21, wu2024marabou, LyKoKoWoLiDa20}. DNN verifiers take as input a DNN and a specification of the desired property and produce either a \emph{proof} that the property always holds, or a \emph{counterexample} demonstrating a case where the property does not hold.
While the scalability of DNN verifiers has improved dramatically in the past decade~\cite{brix2023fourth}, they struggle when applied to \emph{reactive} (e.g., DRL-based) systems with temporal properties which require reasoning about interactions with the environment over time. This is because a naive approach for reasoning about time requires the involved DNN to be \emph{unrolled} (i.e., a copy made for each time step), greatly increasing the complexity of the verification task.

On the other hand, for dynamical systems, a traditional approach for guaranteeing temporal properties has been to use control certificates such as Lyapunov Barrier functions~\cite{li2023survey}. Unfortunately, standard approaches for constructing these functions are not easily applicable to DRL-based dynamical systems.  Recently, however, techniques have been developed for \emph{learning} control certificates.
 We call these \emph{Neural Lyapunov Barrier} (NLB) certificates~\cite{dawson2023safe}.
Although NLB-based approaches have been shown to work for simple, toy examples, these certificates have been, thus far, difficult to learn and verify for real-world systems, which often involve large state spaces with complex dynamics.

In this work, we 
present a novel framework for training and formally verifying NLB-based certificates.
Our framework can verify both \emph{liveness} and \emph{safety} properties of interest, providing \emph{reach-while-avoid} (RWA) guarantees. We use off-the-shelf DNN verifiers and introduce a set of novel techniques to improve scalability, including certificate \emph{filtering} and \emph{composition}.

We demonstrate our approach with a case study targeting a specific challenge problem, in which the goal is to verify a DRL-based spacecraft controller \cite{ravaioli2022safe}. We show that our framework is able to generate verified NLB-based RWA certificates for a range of complex properties.  These include liveness properties (e.g., \emph{will the spacecraft eventually reach its destination?}) and complex non-linear safety properties (e.g., \emph{the spacecraft will never violate a non-linear velocity constraint}), both of which are challenging to verify using existing techniques. 

%
The rest of this paper is organized as follows:
Sec.~\ref{sec:background} gives an overview of relevant background material on property types, DNN verifiers, and NLB certificates. 
Related work is covered in Sec.~\ref{sec:RelatedWork}. In
Sec.~\ref{sec:approach} and~\ref{sec:compositional}, we present our approach, and Sec.~\ref{sec:evaluation} reports the results of our spacecraft case study.\footnote{Code for reproducing experiments is available at: \href{https://github.com/NeuralNetworkVerification/artifact-fmcad24-docking}{github.com/NeuralNetworkVerification/artifact-fmcad24-docking}} Finally, Sec.~\ref{sec:Conclusion} concludes.

%
%

%
\section{Preliminaries}
\label{sec:background}

\subsection{Property Types}

This work focuses on DRL controllers that are invoked over discrete time steps.
We consider both safety and liveness properties~\cite{AlSc87}.
%

\paragraph{Safety.}
A safety property indicates that \emph{a bad state is never reached}. More formally, let $\states$ be the set of system states, and let $\traj \subseteq \states^*$ be the set of possible system trajectories.  The system satisfies a \emph{safety} property $P$ if and only if every state in every trajectory satisfies $P$:
\begin{equation}
\forall\, \alpha:\alpha \in \traj:(\forall\,\state  \in \alpha:\: \state\vDash P)
\end{equation}
A violation of a safety property is a finite trajectory ending in a ``bad'' state (i.e., a state in which $P$ does not hold).

\paragraph{Liveness.}
A liveness property concerns the \emph{eventual} behavior of a system (e.g., \emph{a good state is eventually reached}). 
More formally, we say a \emph{liveness} property $P$ holds if and only if there exists a state $\state$ in every infinite trajectory where $P$ holds.  Letting $\traj^{\infty}$ be the set of infinite trajectories, we can formalize this as follows.
\begin{equation}
\forall\, \alpha :\alpha \in \tau^{\infty}:\:(\exists\, \state \in \alpha:\: \state\vDash P),
\end{equation}
A violation of a liveness property is an infinite trajectory in which each state violates the property $P$.

\subsection{DNNs, DNN Verification, and Dynamical Systems.} 

\paragraph{Deep Learning.} 
Deep neural networks (DNNs)~\cite{GoBeCo16} consist of layers of neurons, each layer performing a (typically nonlinear) transformation of its input. 
This work focuses on deep reinforcement learning (DRL), a popular paradigm in which a DNN is trained to realize a \emph{policy}, i.e., a mapping from states (the DNN's inputs) to actions (the DNN's outputs), which is used to control a reactive system. For more details on DRL, we refer to~\cite{Li17}.

\paragraph{DNN Verification.}
Given \begin{inparaenum}[(i)]
		\item a trained DNN (e.g., a DRL agent) $N$;
		\item a precondition $P$ on the DNN's inputs, limiting the input assignments; and	
		\item a postcondition $Q$ on the DNN's output%
\end{inparaenum},
 the goal of DNN verification is to determine whether the property $P(\state)\rightarrow Q(N(\state))$ holds for any neural network input $\state$. In many DNN verifiers (a.k.a., \emph{verification engines}), this task is equivalently reduced to determining the satisfiability of the formula $P(\state)\land \neg Q(N(\state))$. If the formula is satisfiable (\sat), then there is an input that satisfies the pre-condition and violates the post-condition, which means the property is violated. On the other hand, if the formula is unsatisfiable (\unsat), then the property holds. It has been shown~\cite{KaBaDiJuKo17} that verification of piecewise-linear DNNs is NP-complete.

\paragraph{Discrete Time-Step Dynamical Systems.}
We focus on dynamical systems that operate in a discrete time-step setting.
More formally, these are systems whose trajectories satisfy the equation:
\begin{equation}
    \state_{t+1} = \trans(\state_t, \control_t),
\end{equation}
where $\trans$ is a \emph{transition function} that takes as inputs the current state $\state_t\in\states$ and a control input $\control_t\in\controls$ and produces the next state $\state_{t+1}$. These systems are controlled using a feedback control policy $\ctrlfun:\states \rightarrow \controls$ which, given a state $\state\in\states$ produces control input $\control = \ctrlfun(\state)$. 
In our setting, the controller $\ctrlfun$ is realized by a DNN trained using DRL. DRL-based controllers are potentially useful in many real-world settings, due to their expressivity and their ability to generalize to complex environments \cite{TalSobKir2019}.



\subsection{Control Lyapunov Barrier Functions}

The problem of verifying a liveness or safety property over a dynamical system with a given control policy can be reduced to the task of identifying a certificate function $\certfun: \states\mapsto \mathbb{R}$, whose input-output relation satisfies a particular set of constraints that imply the property. 
There are two fundamental types of certificate functions.

\paragraph{Lyapunov Functions.}
A Lyapunov function (a.k.a. \emph{Control Lyapunov function}) represents the energy level at the current state: 
as time progresses, energy is dissipated until the system reaches the zero-energy equilibrium point~\cite{lyapunovbook}. 
Hence, such functions are typically used to provide 
\textit{asymptotic stability}, i.e., 
adherence to a desired liveness property, or the eventual convergence of the system to some goal state. 
Such guarantees can be afforded
by learning a function that ($i$) reaches a $0$ value at equilibrium, ($ii$) is strictly positive everywhere else; and ($iii$) either monotonically decreases~\cite{NLC, Chang2021} or decreases by a particular constant~\cite{ganai2023learning} with each time step.

\paragraph{Barrier Functions.}
Barrier functions~\cite{cbfsurvey}, a.k.a. \emph{Control Barrier Functions}, 
are also energy-based certificates.  However, these functions are typically used for verifying safety properties. 
Barrier functions enforce that a system will never enter an unsafe region in the state space.  This is done by assigning unsafe states a function value above some threshold and then verifying that barrier function never crosses this threshold~\cite{cbfqptac, ames_robust, basile1969controlled}.

\paragraph{Control Lyapunov Barrier Functions.}
In many real-world settings, it can be useful to verify both liveness properties and safety properties. 
In such cases, a
\emph{Control Lyapunov Barrier Function} (CLBF) can be used, which combines the properties of both Control Barrier functions and Lyapunov functions. 
CLBFs can provide rigorous guarantees w.r.t. a wide variety of temporal properties, including the general setting of reach-while-avoid tasks~\cite{EdwPerAba2023}, which we describe next. 

%
%

\paragraph{Reach-while-Avoid Tasks.}
In a \emph{reach-while-avoid} (RWA) task, we must find a controller $\pi$ for a dynamical system such that all trajectories $\{\state_1, \state_2 ... \}$ produced by this controller ($i$) do not include unsafe (``bad'') states; and ($ii$) eventually reach a goal state. 
More formally the problem can be defined as follows:



\begin{definition}[Reach-while-Avoid Task]
\label{def:reachavoid}
    \textcolor{white}{ }\\
   \textbf{Input: } 
   A dynamical system with a set of initial states $\states_I\subseteq\states$, a set of goal states $\states_G\subseteq \states$, and a set of unsafe states $\states_U\subseteq \states$, where $\states_I\cap \states_U =\emptyset$ and $\states_G \cap \states_U = \emptyset$\\
   \textbf{Output: }
   A controller $\pi$ such that for every trajectory $\traj=\{\state_1, \state_2 ... \}$ satisfying $\state_1 \in \states_I$:
    \begin{enumerate}
        \item \textbf{Reach:} $\exists\, t \in \mathbb{N}.\: x_t \in \states_G$
        \item \textbf{Avoid:} $\forall\, t\in \mathbb{N}.\: x_t\not\in \states_U$ 
    \end{enumerate}
\end{definition}


\section{Related Work}
\label{sec:RelatedWork}

\subsection{Control Certificates}
Control certificate-based approaches form a popular and effective class of methods for providing guarantees about complex dynamical systems in diverse application areas including robotics~\cite{dawson2023safe}, energy management~\cite{huang2021neural}, and biomedical systems~\cite{de2022intelligent}. Control Lyapunov functions are certificates for system stability, and the closely-related control Barrier functions are certificates for safety. While such Lyapunov-based certificates have been proposed over a century ago~\cite{lyapunov1892general}, their main drawback lies in their computational intractability~\cite{giesl2015review}. As a result, practitioners have mainly relied on unscalable methods for constructing certificates, such as manual design for domain-dependent certificate functions~\cite{Choi2020, Castaneda2020}, sum-of-squares approaches restricted to polynomial systems~\cite{Jarvis2003, Majumdar2017}, and quadratic programming~\cite{li2023survey}. 

\subsubsection{Formal Verification of Neural Certificates}
Recent methods have leveraged neural networks as verifiable models of these control certificates, forming a class of \textit{neural certificate} approaches~\cite{dawson2023safe}. For a fixed controller,~\cite{Spencer2018} distills the problem into solving binary classification with neural networks, but the method is limited to polynomial systems and only obtains a region of attraction, making it incompatible with most RWA problems, which have a predefined goal region.

In~\cite{abate2020formal,ahmed2020automated}, SMT solvers are employed to check whether a certificate for a specific controller satisfies the Lyapunov conditions and, if not, to return counterexamples which can be used to retrain the neural certificate.  A similar approach can be used for Barrier conditions~\cite{peruffo2021automated}. In~\cite{abate2021fossil, edwards2023fossil}, the Fossil tool is introduced, which combines these methods. In~\cite{abate2024safe}, Fossil is used to generate training examples for barrier certificates which are used to construct overapproximations of safe reach sets. However, these methods require verifying all constraints in the certificate for the entirety of the relevant state space --- a task which can be computationally prohibitive (as we show in Section~\ref{sec:evaluation}). 

In~\cite{NLC}, a Neural Lyapunov Control (NLC) framework is proposed, which jointly learns the Lyapunov certificate and the controller. The algorithm iteratively calls the dReal SMT solver~\cite{gao2013dreal} to generate counterexamples and retrain both the neural certificate and the control policy. Various extensions and applications followed:~\cite{grande2023augmented} addresses algorithmic problems in NLC;~\cite{grande2023systematic} automates the design of passive fault-tolerant control laws using NLC;~\cite{zhou2022neural} extends NLC to unknown nonlinear systems;~\cite{wu2023neural} extends NLC to discrete-time systems;~\cite{samanipour2023stability} verifies single hidden-layer ReLU neural certificates with enumeration~\cite{rada2018new} and linear programming; and~\cite{zhang2024exact} develops a framework for Barrier functions when there is an existing nominal controller. However, these methods do not consider the more general reach-avoid problem.

\subsubsection{Data-driven Neural Certificates}
\label{sec:DDNC}

To improve scalability, a recent line of research proposes learning certificates and controllers from online and/or offline data without additional formal verification~\cite{dawson2023safe}, following the intuition that, with increasing data, the number of violations in the trained certificate will tend toward zero~\cite{boffi2021learning}. ~\cite{Chang2021,ganai2023learning} learn Lyapunov certificates for stabilization control, and~\cite{qin2022quantifying, qin2021learning, tong2023enforcing, yu2023sequential} synthesize neural Barrier functions in various settings like multi-agent control, neural radiance field~\cite{mildenhall2021nerf} imagery, and pedestrian avoidance. These methods (by design) cannot provide rigorous guarantees on the validity of their learned certificates.


\subsection{Reach-Avoid methods} 

Solutions for tasks requiring the simultaneous verification of both liveness and safety properties, of which the RWA task is a common example, have also relied on control theoretic principles.~\cite{dawson2022safe} learns a combined Lyapunov and Barrier certificate to construct controllers with stabilization and safety guarantees. The Hamilton-Jacobi (HJ) reachability-based method (a verification method for ensuring optimal control performance and safety in dynamical systems~\cite{hjreachabilityoverview}) has also been used to solve reach-avoid problems~\cite{fisac2015reach,hsu2021safety,so2023solving}. Safe reinforcement learning is closely related to reach-avoid: the goal is to maximize cumulative rewards while minimizing costs along a trajectory~\cite{brunke2022safe}, and it has been solved with both Lyapunov/Barrier methods~\cite{yang2023model, Chow2018} and HJ reachability methods~\cite{yu2022reachability, ganai2023iterative}.  As mentioned, scalability is a crucial challenge in this context.  The next section describes our approach for addressing this challenge.





\section{Reach-While-Avoid Certificates}
\label{sec:approach}


In this section, we present our approach for scalably creating verified NLB certificates.
We first describe reach-while-avoid (RWA) certificates, a popular class of existing NLB-based certificates. We next present an extension called \emph{Filtered} RWA certificates, which significantly simplifies the learning task and enables efficient training of certificates for complex properties. We then present a \emph{compositional} certification approach, which independently trains a series of certificates that can be jointly verified to handle even larger state spaces.



\subsection{RWA certificates}
A function $\certfun:\states\mapsto \mathbb{R}$ is an RWA certificate for the Reach-Avoid task in Definition~\ref{def:reachavoid} if, for some $\alpha > \beta$ and $\epsilon > 0$, it satisfies the following constraints.%
\footnote{These constraints are similar to the ones defined in prior work~\cite{EdwPerAba2023} but are specific to discrete time-step systems and instead place constraints on the set of unsafe states instead of a compact safe set.}
\begin{align}
& \forall\,\state\in\states_I. && \certfun(\state) \le \beta
\label{eq:CLBFcond1} \\
& \forall\,\state\in\states\setminus\states_G. && \certfun(\state) \le \beta \rightarrow \certfun(\state) - \certfun(\trans(\state,\ctrlfun(\state))) \geq \epsilon 
\label{eq:CLBFcond2} \\
& \forall\,\state\in\states_U. && \certfun(\state) \ge \alpha \label{eq:CLBFcond3}
\end{align}
Any tuple of values $(\alpha,\beta,\epsilon)$ for which these conditions hold is called a \emph{witness for} the certificate.
RWA certificates provide the following guarantees.\footnote{The proof can be found in Appendix~\ref{sec:appendix:proofs}.}
\begin{lemma}
If \certfun is an RWA certificate for a dynamical system with witness $(\alpha,\beta,\epsilon)$, and $V$ has a lower bound,\footnote{This is always the case if the output of $V$ is implemented using a finite representation such as floating-point arithmetic.} then for every infinite trajectory $\traj$ starting from a state $\state\in\states\setminus\states_G$ such that $\certfun(\state)\le \beta$, $\traj$ will eventually contain a state in $\states_G$ without ever passing through a state in $\states_U$.
\end{lemma}

\noindent
Intuitively, $\certfun$ partitions the state space into three regions:
\begin{itemize}
    \item a \emph{safe region} where the value of the certificate is at most $\beta$. This region includes the initial states $\states_I$ and any states reachable from $\states_I$. 
    Furthermore, starting from any non-goal state in the safe region, the certificate function value should decrease by \emph{at least} $\epsilon$ at each time step.
    \item an \emph{unsafe region} where the value of the certificate is at least $\alpha$. This region must include the unsafe states $\states_U$.
    \item an \emph{intermediate region}, where the value of the certificate is strictly between $\beta$ and $\alpha$. States in this region are not unsafe but are also not reachable from $\states_I$. This can also be thought of as a ``buffer'' region that separates the safe region from the unsafe region.  These states play a role in the compositional approach described below.
\end{itemize}




\subsection{FRWA certificates}

A \emph{neural RWA} certificate is an RWA certificate realized by a DNN.  Such a DNN can be trained by following the NLC approach~\cite{NLC}, using the constraints ~\eqref{eq:CLBFcond1}--\eqref{eq:CLBFcond3} as training objectives.  Because we are also interested in formally verifying these certificates, we would like to keep the DNNs (both the controller and the certificate) small so that verification remains tractable.  We have observed that this can be challenging when the system and properties are non-trivial.  To help address this, we introduce 
an improvement called \emph{Filtered Reach-while-Avoid} (FRWA) certificates.



The idea behind FRWA is straightforward.  Often, we can describe the goal and unsafe regions using simple predicates (or filters) on the state space.
We pick constants $c_1, c_2$ such that $c_1 \le \beta < \alpha \le c_2$ and then hard-code the implementation of $\certfun$ so that $\state\in\states_G \to \certfun(\state) = c_1$ and $\state\in\states_U \to \certfun(\state) = c_2$.
Note that the latter ensures that condition~\eqref{eq:CLBFcond3} holds by construction.
Importantly, this not only makes the training task easier, but also reduces the number of queries required to formally verify the certificate. On the other hand, hard-coding the certificate value for inputs in $\states_G$ makes it easier to learn constraint \eqref{eq:CLBFcond2}. The reason for this is more nuanced. If we randomly initialize the certificate neural network, the certificate value for some states in $\states_G$ could start out larger than $\beta$, making it more difficult to satisfy constraint~\eqref{eq:CLBFcond2} for a point $\state$ where $\certfun(\state) \le \beta$ and $\trans(\state,\ctrlfun(\state)) \in \states_G$. Fixing the certificate values for states in $\states_G$ to at most $\beta$ (ideally, significantly below $\beta$) ensures that, at least for such points, condition~\eqref{eq:CLBFcond2} is easier to satisfy. In practice, FRWA certificates can be implemented by using a wrapper around a DNN which checks the two filters and only calls the DNN if they both fail.
The practical effectiveness of FRWA certificates is demonstrated in Sec.~\ref{sec:evaluation}. 

\paragraph{FRWA Training.}
FRWA simplifies the certificate learning process, as now, only constraints \ref{eq:CLBFcond1} and \ref{eq:CLBFcond2} are relevant for training.  We custom design the reinforcement learning training objective function as follows.  Let $\state_1,\dots,\state_N$ be the set of training points, and let $\state'_i=\trans(\state_i,\ctrlfun(\state_i))$.  We define:

\begin{align}
O_s &= c_s \sum_{i \,|\, \state_i \in \states_I} \frac{\text{ReLU}(\delta_1 + \certfun(\state_i) - \beta)}{\sum_{i \,|\, \state_i \in \states_I} 1} \label{eq:ReluOne} \\
O_d &= c_d \!\!\!\!\!\!\sum_{i \,|\, \state_i \in \states \setminus (\states_U \cup \states_G), \certfun(\state_i) \le \beta} \frac{\text{ReLU}(\delta_2 + \epsilon + \certfun(\state'_i) - \certfun(\state_i))}{\sum_{i \,|\, \state_i \in \states \setminus (\states_U \cup \states_G), V(\state_i) \le \beta} 1} \label{eq:ReluTwo} \\
O &= O_s + O_d \label{eq:ReluFour}
\end{align}


Eq.~\eqref{eq:ReluOne} penalizes deviations from constraint~\eqref{eq:CLBFcond1}, and Eq.~\eqref{eq:ReluTwo} penalizes deviations from constraint~\eqref{eq:CLBFcond2}.
We incorporate parameters $\delta_1 > 0$ and $\delta_2 >0$, which can be used to tune how strongly the certificate over-approximates adherence to each constraint.
Similarly, constants $c_s$ and $c_d$ can be used to tune the relative weight of the two objectives. The final training objective $O$ in \eqref{eq:ReluFour} is what the optimizer seeks to minimize, by using stochastic gradient descent (SGD) or other optimization techniques. 
We note that the FRWA certificates are trained in a self-supervised, non-RL setting.





\paragraph{FRWA Data Sampling.}
From the formulation above, we see that
only data points in  $(\states \setminus (\states_U \cup \states_G)) \cup \states_I$ affect the objectives, and thus, only these data points need to be sampled.

%
%

\paragraph{FRWA Verification.}
We use DNN verification tools to formally verify that conditions \eqref{eq:CLBFcond1}-\eqref{eq:CLBFcond3} hold for our certificates.  Filtering introduces a slight complication.  Recall that a FRWA certificate is implemented as a wrapper around a DNN, meaning that the DNN itself can behave arbitrarily when
either $\state \in \states_G$ or $\state \in \states_U$.
Fortunately, we can adjust the verification conditions for the DNN part of the certificate as follows.

Constraint~\eqref{eq:CLBFcond1} can be checked as is.  The filtering does not affect this property.  And it is easy to see that checking the property for the DNN does indeed ensure the property holds for the full certificate.

Constraint~\eqref{eq:CLBFcond3} need not be checked at all, as the filtered certificate ensures this condition by construction.

Verification of constraint \ref{eq:CLBFcond2} is done by instead checking that: 
\begin{align}
& \forall\, \state \in \states \setminus (\states_U \cup \states_G), \state' \in \states. \nonumber \\
& \quad (\state' = \trans(\state,\ctrlfun(\state)) \wedge \certfun(\state) \le \beta) \to \nonumber \\
& \quad\quad (\certfun(\state) - \certfun(\state') \ge \epsilon \vee (\state' \in \states_G)) \wedge (\state' \not\in \states_U)\label{eq:CLBFcond2mod}
\end{align}

\noindent
There are three main differences between~\eqref{eq:CLBFcond2} and \eqref{eq:CLBFcond2mod}.  Since the filter ensures that $\certfun(\state)>\beta$ when $\state\in\states_U$, we can safely exclude states in $\states_U$ from the check.  Similarly, if the system ever transitions from a state $\state$ with $\certfun(\state) \le \beta$ to an unsafe state, the filter ensures that condition~\eqref{eq:CLBFcond2} is violated, so it suffices to check that $\state'\not\in\states_U$ to cover this case.

The last difference is a bit more subtle.  Observe that \eqref{eq:CLBFcond2mod} is trivially true if $\state'\in\states_G$, meaning that if we transition to a goal state, we do not enforce \eqref{eq:CLBFcond2}. 
 However, it is easy to see that Lemma~\ref{lem:1} still holds with this relaxed condition: if every transition either reduces $\certfun$ by at least $\epsilon$ or reaches a goal state, then clearly, we must eventually reach a goal state.

{\subsection{CEGIS loop}
We use a \emph{counterexample-guided inductive synthesis} (CEGIS) loop, shown in Fig.~\ref{fig:cegis}, to obtain a fully verified controller and certificate.
We first train an initial controller $\ctrlfun$. 
Then, at each CEGIS iteration, we jointly train $\certfun$ and $\ctrlfun$ until a loss of $0$ is obtained and then use a sound and complete DNN verifier (we use \emph{Marabou}~\cite{KaHuIbJuLaLiShThWuZeDiKoBa19} in our experiments) to identify counterexamples. 
If the verifier identifies a counterexample violating constraints~\eqref{eq:CLBFcond1} or \eqref{eq:CLBFcond2} (recall that constraint \eqref{eq:CLBFcond3} is satisfied by construction), we sample points in the proximity of the counterexample and use these to augment the training data. By sampling multiple nearby points, we hope to influence the training to learn smooth behavior for a localized neighborhood instead of overfitting to a specific point.
This process is repeated iteratively until no counterexamples are found, at which point we are guaranteed to have produced a fully verified controller and certificate.

\begin{figure}[!t]
  \centering
  \includegraphics[width=0.8\columnwidth]{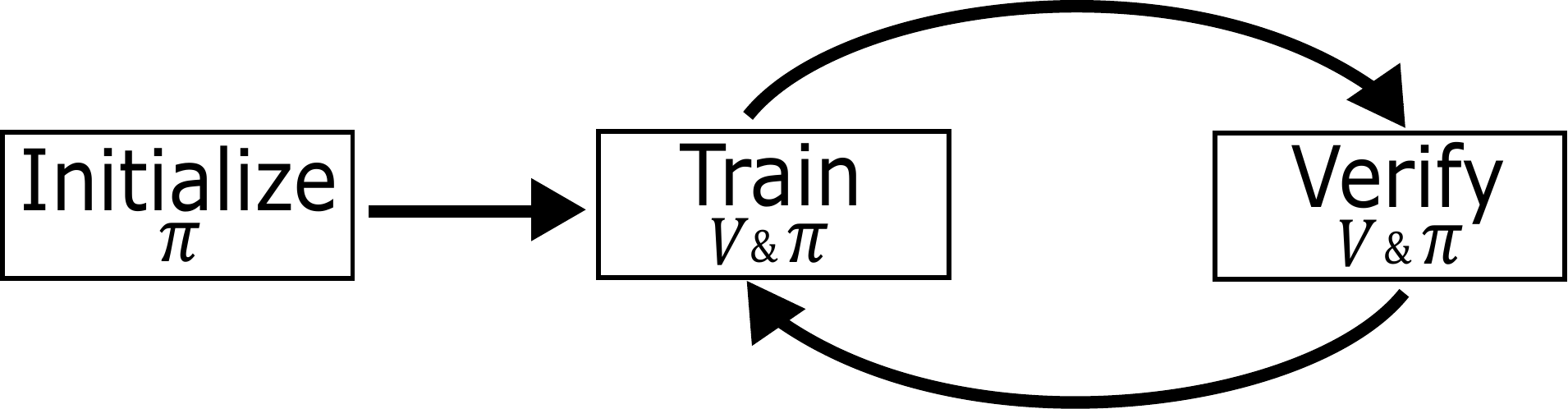}
  \caption{The CEGIS Loop used to iteratively train and verify controller $\ctrlfun$ and certificate $V$. Verification counterexamples are used to augment the training dataset.}
  \label{fig:cegis}
\end{figure}

\section{Compositional Certificates}
\label{sec:compositional}

While filtering does improve the efficiency of both training and verification, the approach outlined above still suffers from scalability challenges, especially as the system complexity or state space covered by the controller increases.  In this section, we introduce \emph{compositional certificates}, which aim to aid scalability by training multiple controller-certificate pairs, each covering different parts of the state space.
%
The certificates are compositional in the sense that a simple meta-controller can be designed to determine which controller-certificate pair to use when in a given state, and we can formally guarantee that the meta-controller satisfies the requirements of definition~\ref{def:reachavoid}.


\paragraph{CRWA.} Formally, a compositional RWA certificate (CRWA) for an RWA task is composed of $n$ RWA certificates,\footnote{Each controller in a CRWA can make use of the FRWA technique described above.} which we denote $\certfun_0,\dots,\certfun_{n-1}$, with corresponding controllers, which we denote $\ctrlfun_0,\dots,\ctrlfun_{n-1}$, with $n \ge 2$.  Furthermore, each pair $(\certfun_i,\ctrlfun_i)$ must be an RWA certificate with some witness $(\alpha_i,\beta_i,\epsilon_i)$ for an RWA task whose dynamics are that of the main RWA task, but whose parameters are $(\states^i_I,\states^i_G,\states^i_U)$.  These parameters must satisfy the following conditions:
\begin{itemize}
\item[($i$)] $\states^0_I \subseteq \states_I$, $\states^0_G = \states_G$, and $\states_U \subseteq \states^0_U \subseteq \overline{(\states^0_I \cup \states^0_G)}$, where $\overline{S}$ denotes the complement of the set $S$;
\item[($ii$)] for $0 < i < n$, $\states^{i-1}_I \subseteq \states^i_I \subseteq \states_I$, $\states^{i}_G = \{x \in \overline{\states^{i-1}_U} \ \mid\ \certfun_{i-1}(x) \le \beta_{i-1} \} \cup \states^0_G$, and $\states_U \subseteq \states^i_U \subseteq \states^{i-1}_U$; 
\item[($iii$)] either $\states^i_I \neq \states^{i-1}_I$ or $\states^i_U \neq \states^{i-1}_U$; and
\item[($iv$)] $\states^{n-1}_I = \states_I$ and $\states^{n-1}_U = \states_U$.
\end{itemize}
Intuitively, the idea is as follows.  We start with an initial controller capable of guiding the system from some \emph{subset} of the initial states $\states_I$ to the original goal states $\states_G$ while avoiding some \emph{superset} of the unsafe states $\states_U$.  Then, for each subsequent controller, we ensure that it can guide the system either from a larger subset of the initial states $\states_I$ or while avoiding a smaller superset of the unsafe states $\states_U$, or both, to a new goal region consisting of the states considered safe by the previous controller, i.e., the states $x$ for which $\certfun(x)\le \beta$.  For the final controller (controller $n-1$), the set of initial and unsafe states should coincide with those of the original RWA problem.  Note that the algorithm does not say how to choose of $\states_I^i$ and $\states_U^i$ for $i<n-1$ other than to specify that these sets should be monotonically increasing and decreasing, respectively.  Finding good heuristics for choosing these sets in the general case is a promising direction for future work.

The meta-controller behaves as follows.  Given any starting state $x\in\states_I$, we first check if $x\in\states_G$.  If so, we are done. 
 Otherwise, we determine the smallest $i$ for which $x\in\states^{i}_I$ and guide the system using $\ctrlfun_i$ until a state in $\states^i_G$ is reached, which will occur in some finite number of steps because of the guarantees provided by $\certfun_i$.  At this point, we transition to $\ctrlfun_{i-1}$, and the process repeats until a state in $\states_G$ is reached.

\begin{algorithm}[t]
\caption{CRWA Training and Verification}
\label{alg:comptraining}
\SetKwInOut{Input}{Input}
\SetKwInOut{Output}{Output}
\SetKwRepeat{Do}{do}{while}
\Input{$\states_I$, $\states_G$, $\states_U$}
\Output{$\ctrlfun_0,\dots,\ctrlfun_{n-1},\certfun_0,\dots,\certfun_{n-1}$ for some $n$}
 $\states^0_G \gets \states_G$\\
 Choose $\states^0_I \subseteq \states_I$ and $\states^0_U \supseteq \states_U$, with $\states^0_U \subseteq \overline{(\states^0_I \cup \states^0_G)}$\\
 Choose $\alpha^0 > \beta^0$ and $\epsilon^0 > 0$\\
 Train and verify controller $\pi_0$ and certificate $V_0$ with witness $(\alpha^0,\beta^0,\epsilon^0)$ for the RWA task corresponding to $\states^0_I,\states^0_G,$ and $\states^0_U$ using, e.g., the approach shown in Fig. \ref{fig:cegis}\\
 $i \leftarrow 0$\\
\While{$\states^i_I \subset \states_I$ OR $\states^i_U \supset \states_U$}{
    $i \leftarrow i + 1$\\
    Choose $\states^i_I$, $\states^i_U$ such that $\states^i_I \supset \states^{i-1}_I$ or $\states^i_U \subset \states^{i-1}_U$\\
    $\states^i_G \gets \{x \in \overline{\states^{i-1}_U}\ \mid\ \certfun_{i-1}(x) \le \beta_{i-1}\} \cup \states^0_G$ \\
    Choose $\alpha^i > \beta^i$ and $\epsilon^i > 0$\\
    Train and verify controller $\ctrlfun_i$ and certificate $\certfun_i$ with witness $(\alpha^i,\beta^i,\epsilon^i)$ for the RWA task corresponding to $\states^i_I,\states^i_G,$ and $\states^i_U$ using, e.g., the approach shown in Fig. \ref{fig:cegis}\\
}
\end{algorithm}

\begin{figure}[!t]
\label{fig:compositional}
  \centering
  \includegraphics[width=\columnwidth]{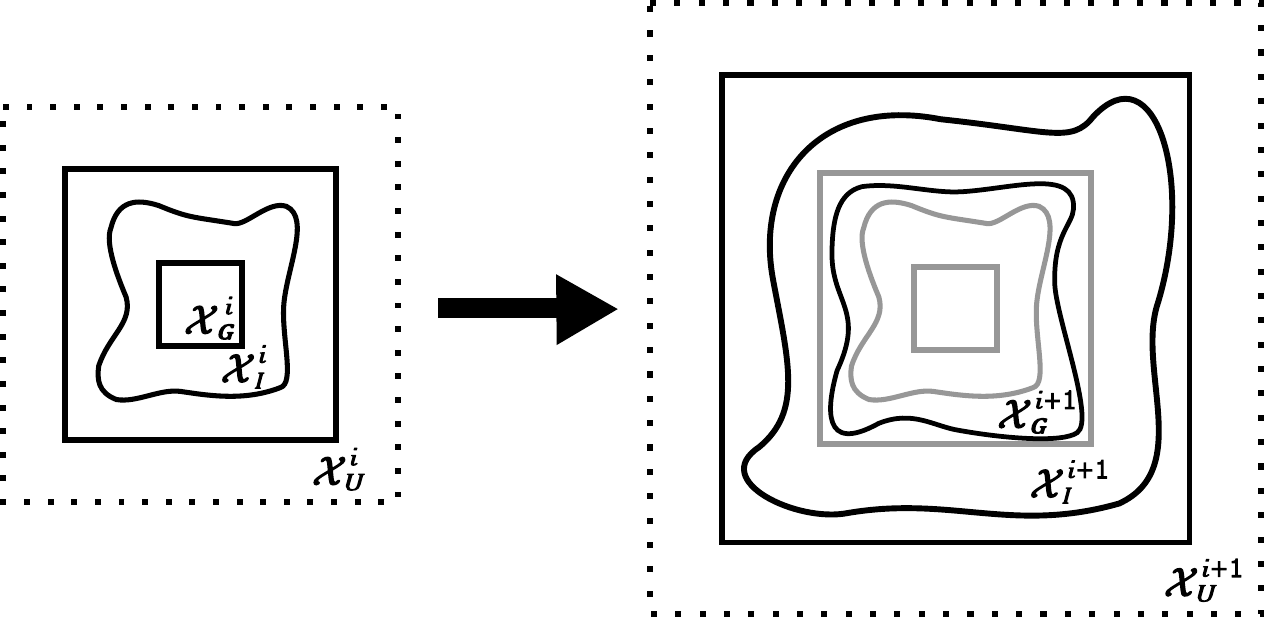}
  \caption{Visualization of how consecutive certificates relate when building a CRWA certificate. Note that $\states_G$ need not be a subset of $\states_I$. The dotted lines indicate that the unsafe state region extends infinitely outside the solid line box.  Wavy lines indicate outer boundaries for initial or goal regions.}
  \label{fig:sub_regions}
\end{figure}

The training and verification of a CRWA certificate is described in Alg.~\ref{alg:comptraining} and visualized in Fig.~\ref{fig:sub_regions}.

%
The following lemma captures the correctness of our approach.%
\footnote{Due to space constraints, the proof of this lemma is in Appendix \ref{sec:appendix:proofs}.}

\begin{lemma}
{Given a CRWA certificate for an RWA task with parameters $\states_I$, $\states_G$, and $\states_U$, all trajectories guided by the meta-controller starting at any point in $\states_I$ will reach $\states_G$ in a finite number of steps while avoiding $\states_U$. In other words, a CRWA certificate provides a correct solution for the RWA task.}
\end{lemma}

\paragraph{CRWA Data Sampling.} When training certificate $V_i$, it is important that the training dataset contains sufficient states sampled from $\states^i_I \setminus \states^i_G$. Otherwise, $V_i$ might learn to assign values greater than $\beta^i$ as much as possible in order to meet constraint \eqref{eq:CLBFcond2}, as opposed to appropriately assigning all states in $\states^i_I\setminus\states^i_G$ to have values less than $\beta^i$, due to an insufficient loss penalty for constraint \eqref{eq:CLBFcond3}. To ensure that states in the region $\states^i_I \setminus \states^i_G$ are included in the training data, we can identify states over constrained subspaces in $\states^i_I \setminus \states^i_G$, and then include in the data set those points as well as a random subset of their neighbors which likely lie in the same region.



\paragraph{Tradeoffs in choosing Intermediate Goals for CRWA certificates.} 
%
It is possible to further reduce the state space for individual certificates in a CRWA certificate by using a more precise description of the goal states. 
 In particular, we could set the goal states as follows: 
\begin{multline}
    \states^i_G = \{x \in \overline{\states^{i-1}_U}\ \mid\ \certfun_{i-1}(x) \leq \beta_{i-1}\} \cup \states^{i-1}_G.
    \label{eq:CFRWAeq2}
\end{multline}

However, using \ref{eq:CFRWAeq2} leads to a linear increase in the number of DNNs that must be included during training and verification at each iteration of Alg. \ref{alg:comptraining}.  This quickly becomes prohibitively expensive, especially for the verification step.  We thus use the simpler formulation described above. 


\section{Evaluation}
\label{sec:evaluation}

\subsection{Case Study}
\label{subsec:casestudy}

We evaluate our approach on the 2D docking task from~\cite{ravaioli2022safe},\footnote{Additional details on this case study are described in our recent related paper~\cite{Dasc2024Paper}} in which a spacecraft is trained using DRL to navigate to a goal.
More specifically, a DRL agent maneuvers a \emph{deputy} spacecraft, controlled with thrusters that provide forces in the $x$ and $y$ directions. The deputy spacecraft attempts to safely navigate until it reaches a state that is in close proximity to a designated \emph{chief} spacecraft, while obeying a distance-dependent safety constraint.\footnote{Further details are in Appendix~\ref{sec:appendix:DnnTraining}.}
We focus on this benchmark for several reasons:
(i) it has been proposed and studied as a challenge problem in the literature~\cite{ravaioli2022safe},
(ii) there exist natural safety and liveness properties for it; and
(iii) existing approaches have been been unable to formally verify these properties.


%

\paragraph{System Dynamics.}
The system is modeled using the Clohessy-Wiltshire relative orbital motion linear approximation in the non-inertial Hill's reference frame, with the \emph{chief} spacecraft lying at the origin \cite{clohessy1960terminal, hill1878researches}. The state of the system, $\boldsymbol{\state} = [x, y, \dot{x}, \dot{y}]^T$, includes the position in $(x,y)$ and the velocities in each direction, $(\dot{x},\dot{y})$.  The control input is $\boldsymbol{\control} = [F_x, F_y]$, where $F_x$ and $F_y$ are the thrust forces applied along the $x$ and $y$ directions, respetively. Each thrust force component is allowed to range between $-1$ and $+1$ Newtons (enforced with standard piecewise linear clipping). As in the original scenario~\cite{ravaioli2022safe}, the spacecraft's mass, $m$, is 12kg.
The continuous time state dynamics of the system are determined by the following ordinary differential equations (ODE), with $n = 0.001027$ rad/s:

\begin{align}
\dot{\boldsymbol{\state}} &= [\dot{x}, \dot{y}, \ddot{x}, \ddot{y}]^T \\
\ddot{x} &=  2n\dot{y} + 3n^2x + \frac{F_x}{m}\\
\ddot{y} &= -2n\dot{x} + \frac{F_y}{m}
\end{align}

This, in turn, is converted to a discrete system (with a time-step of $T$) by numerically integrating the continuous time dynamics ODE:
\begin{equation}
\boldsymbol{\state}(t_i+T) = \boldsymbol{\state}(t_i) + \int_{t_i}^{t_i + T} \dot{\boldsymbol{\state}}(\tau) d\tau
\end{equation}
The discrete-time version has a closed-form solution that we use to generate successive states for the spacecraft.\footnote{The closed form equations are shown in Appendix \ref{sec:appendix:discretecomp}}. 

\paragraph{Constraints and Terminal Conditions.}
To maintain safety, a distance-dependent constraint is imposed on the \emph{deputy} spacecraft's maximal velocity magnitude (while approaching the \emph{chief}): 
\begin{equation}
\label{eq:safe_constraint}
\sqrt{\dot{x}^2 + \dot{y}^2} \le 0.2 + 2n \sqrt{x^2 + y^2}
\end{equation}
We construct a linear over-approximation of this safety constraints (with OVERT~\cite{SiMaIrKo22}).  It can then be incorporated into the description of the unsafe region.%
\footnote{For additional details, see Appendix \ref{sec:appendix:approx}.}


The goal is for the deputy to successfully dock with the chief without ever violating the velocity safety constraint. 
In the original benchmark~\cite{ravaioli2022safe}, the docking (goal) region is defined as a circle of diameter $d=1$m centered at the origin $(0,0)$. In our evaluation, we use this same goal region for the initial training of the DRL controller.  However, during the CEGIS iteration (i.e., the ``Train'' and ``Verify'' steps in Fig.~\ref{fig:cegis}), we use a conservative subset of this goal region, namely a square centered at the origin whose sides have length $l=0.7$m.  The reason for this is so that the goal region can easily be described using linear inequalities.


\subsection{DNN architecture}

\begin{figure}[t]
  \centering
\includegraphics[width=0.7\columnwidth]{
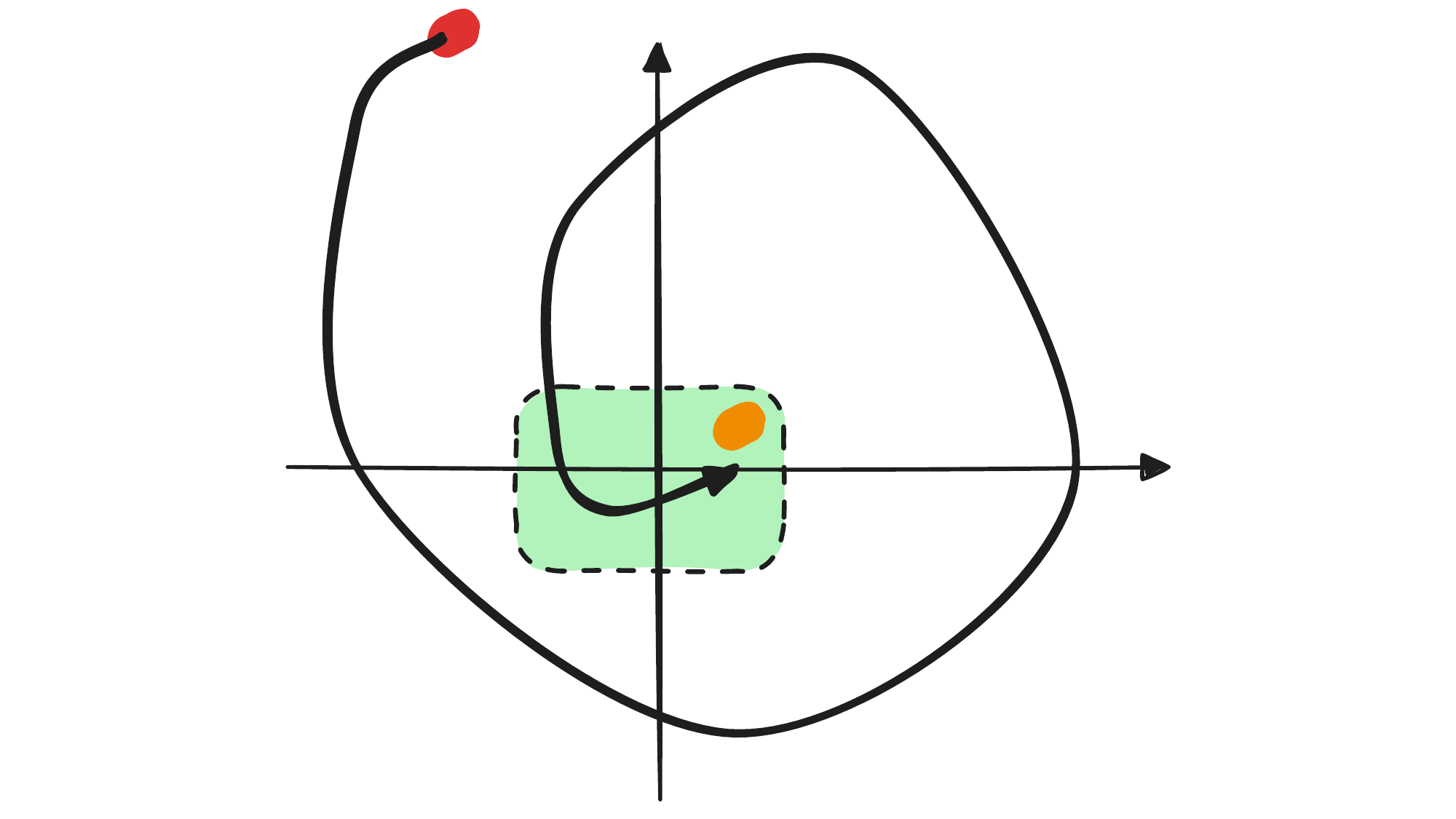}
  \caption{A spiral trajectory: The DRL-controlled spacecraft (starting from the \textcolor{red}{red} point) eventually reaches the destination (\textcolor{orange}{orange}) point within the docking region.}
  \label{fig:spiralDocking}
\end{figure}

We explored various architectures for the DNN used to control the (deputy) spacecraft.  During this exploration phase, we trained each DNN architecture using the original Proximal Policy Optimization RL algorithm implemented by Ray RLlib as described in~\cite{ravaioli2022safe} (without any CEGIS iteration).

After training, we simulated each architecture on 4,000 
random trajectories.  Some selected results are shown in Table~\ref{tab:nn_architecture}.  There are two main observations to take away from these results: $(i)$ while a robust docking capability can be achieved fairly easily, even for small architectures, safety is more difficult and appears to not be robust, even for large architectures; $(ii)$
in all cases, it takes an average of at least 50 steps to dock.
%
The first observation suggests that verification of the liveness (docking) property should be feasible and that training a controller that verifiably achieves both safety and liveness is challenging.  The second observation suggests that
even state-of-the-art DNN verifiers are unlikely to be unable to fully verify the liveness property using the naive unrolling approach~\cite{AmCoYeMaHaFaKa23}. 

We also note that the spacecraft often exhibits highly nonlinear spiral trajectories (as depicted in Fig.~\ref{fig:spiralDocking}), making DNN verification based on induction difficult, as it is difficult to find an inductive property over such irregular trajectories. 
These results help motivate the use of NLB certificates for formal verification of the desired properties.
For the experiments below, we settled on a DNN architecture of two hidden layers with 20 neurons each, and a certificate architecture of two hidden layers with 30 neurons each. Both DNNs use ReLU activations for all hidden layers.  The DNN sizes were chosen based on experimentation and the rough criterion that we wanted the smallest DNNs for which the CEGIS loop would converge in a reasonable amount of time.




\vspace{0.5cm}
\begin{table}[h!]
\centering
\caption{Performance of various DNN architectures. Statistics are collected (per architecture) over $4,000$ trials, with a maximum trajectory length of $2,000$, initialized arbitrarily to set $x,y \in [-10,10]$, but outside the docking region, and $\dot{x} = \dot{y} = 0$.  The first column indicates the number of neurons per hidden layer.}
\label{tab:nn_architecture}
\scalebox{0.9}{
\begin{tabular}{|cccc|}
\hline
DNN Architecture & Safety Success & Docking Success & Average Docking Steps \\
\hline
[4,4] & 100 & 10 & 1,821 \\
\hline
[8,8] & 11 & 100 & 389 \\
\hline
[16,16] & 30 & 100 & 50 \\
\hline
[32,32] & 5 & 100 & 59 \\
\hline
[64,64] & 100 & 100 & 55 \\
\hline
[64,64,64,64] & 100 & 99 & 58 \\
\hline
[200,200] & 92 & 100 & 51 \\
\hline
\end{tabular}
}
\end{table}

\subsection{Implementation and Setup}

The training and verification of the DRL controllers and certificates were carried out on a cluster of Intel Xeon E5-2637 machines, with eight cores of v4 CPUs, running the Ubuntu 20.04 operating system. 
Verification queries were dispatched using the \emph{Marabou} DNN verifier~\cite{KaHuIbJuLaLiShThWuZeDiKoBa19,wu2024marabou} (used in previous DNN safety research~\cite{LaKa21, ReKa21, AmWuBaKa21, AmScKa21, AmZeKaSc22, CaKoDaKoKaAmRe22,AmMaZeKaSc23, AmFrKaMaRe23, BaAmCoReKa23, CoAmKaFa24}) as well as its Gurobi back end.
%
%

For training and verification of RWA, FRWA, and CRWA certificates, we use the following parameters: $\alpha=1+10^{-5}$, $\beta=1$, $\epsilon=10^{-7}$ (the same for all certificates); $c_1 = -10$, $c_2 = 1.2$, $\delta_1 = 10^{-4} - 10^{-5}$, and $\delta_2 = 10^{-4} - 10^{-7}$. These values were determined to work well experimentally.

For weighting of the training objectives, we use $c_s=1$ and $c_d=10$.  The rationale for this is that constraint \eqref{eq:CLBFcond1} is much easier to satisfy than \eqref{eq:CLBFcond2}, so we use the weights to force the 
training to focus on \eqref{eq:CLBFcond2}.

In the CEGIS loop, a learning rate of $5 \times 10^{-3}$ is used to train the first network iteration in the CEGIS loop, and for retraining, a learning rate of $10^{-4}$ is used, since we treat the incorporation of counterexamples as a ``fine-tuning" step and do not want to overfit to the counterexamples.  In the CEGIS loop, we train until a loss of 0 is achieved and then use the verification step to find counterexamples.  We repeat this until there are no more counterexamples or a timeout (12 hours) is reached.



All of our experiments aim to solve RWA tasks, as defined in \cref{def:reachavoid}.  The system dynamics are those of the 2D spacecraft, as described in \cref{subsec:casestudy}.  RWA tasks are parameterized by $\states_G$, $\states_I$, and $\states_U$.  For these sets of states, we typically use square regions centered at the origin.  For convenience, we refer to the set $\{(x,y)\ \mid\ x,y\in[-a,a]\}$ with the abbreviation $[-a,a]$.  For example, as outlined in \cref{subsec:casestudy}, we set $\states_G=[-0.35,0.35]$.  We use different values for $\states_I$, depending on the experiment (indeed, this is the primary variable we vary in our experiments), but whenever $\states_I=[-a,a]$, we then set $\overline{\states_U}=[-(a+1),a+1]$.


\subsection{Experimental Results}

\begin{figure}[t]
\centering
\begin{subfigure}{0.15\textwidth} \includegraphics[width=\linewidth]{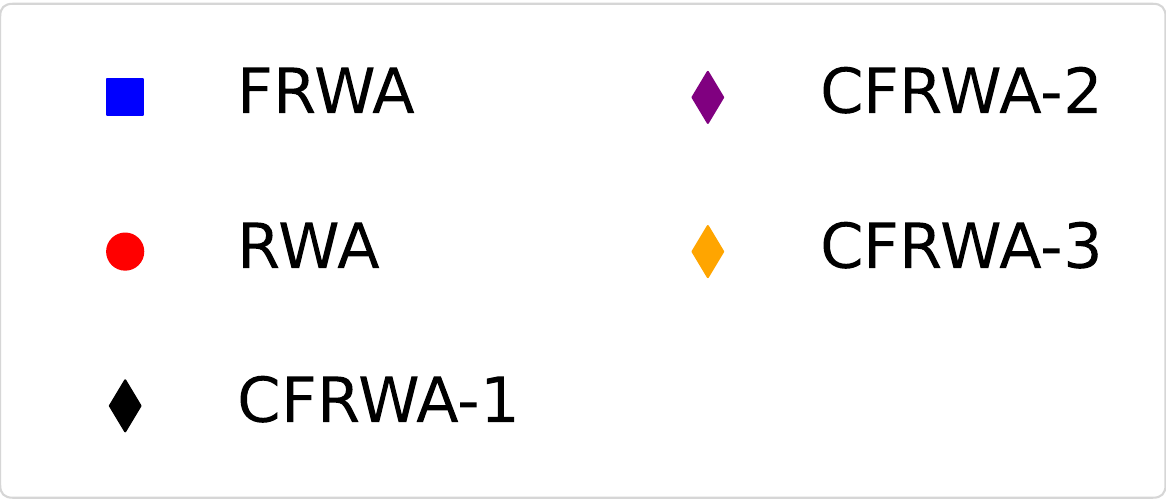} 
    \label{fig:label}
\end{subfigure}
\hfill
\begin{subfigure}{0.15\textwidth}
    \includegraphics[width=\linewidth]{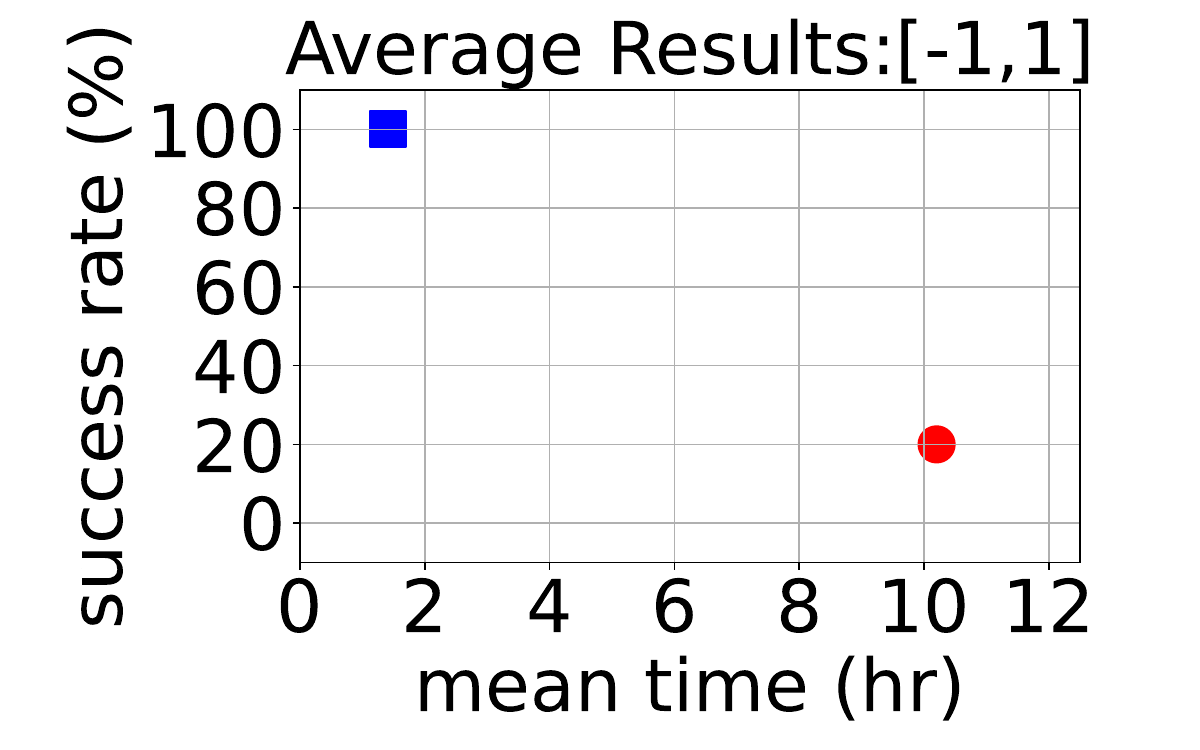} 
    \label{fig:1a}
\end{subfigure}
\hfill
\begin{subfigure}{0.15\textwidth}
\includegraphics[width=\linewidth]{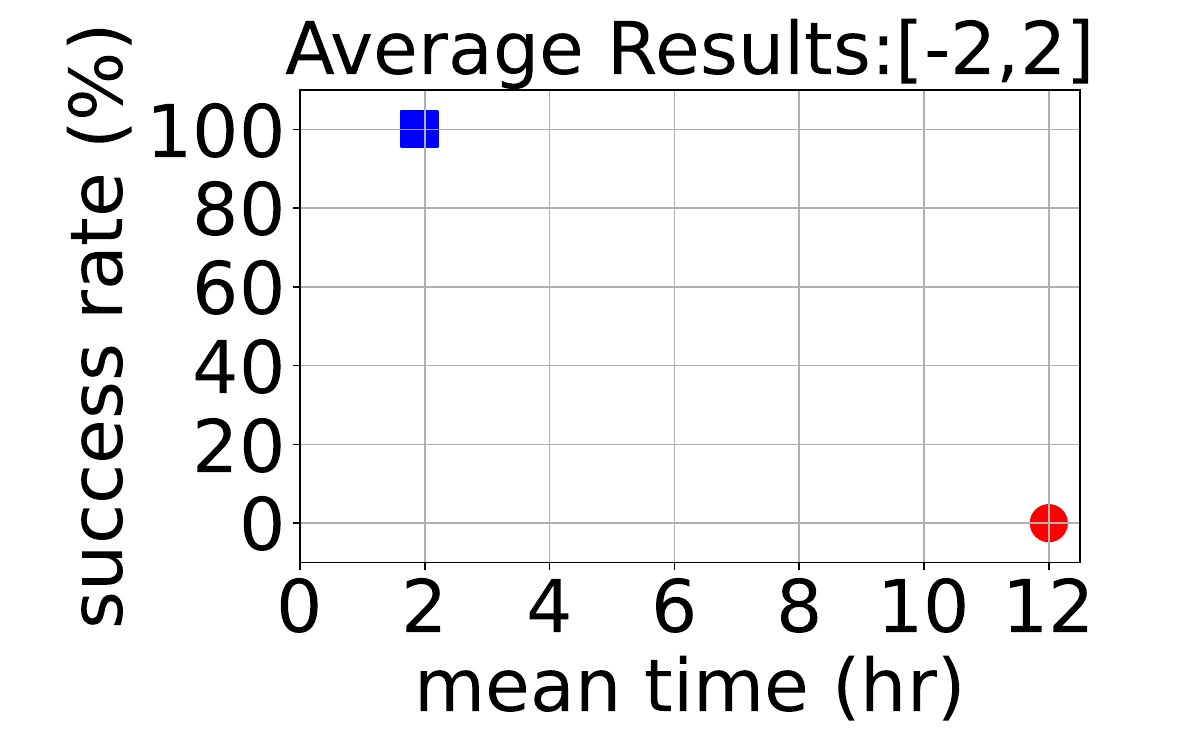}
    \label{fig:1b}
\end{subfigure}
\hfill
\begin{subfigure}{0.15\textwidth}
    \includegraphics[width=\linewidth]{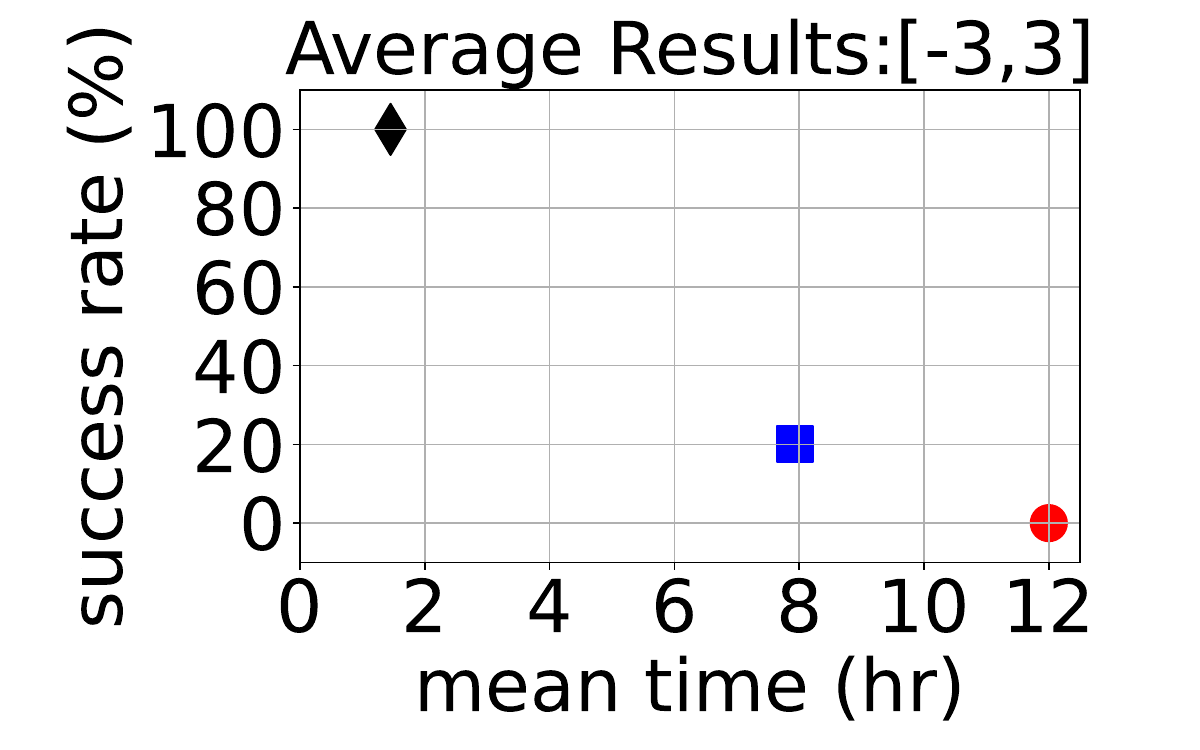}
    \label{fig:1c}
\end{subfigure}
\hfill
\begin{subfigure}{0.15\textwidth}
    \includegraphics[width=\linewidth]{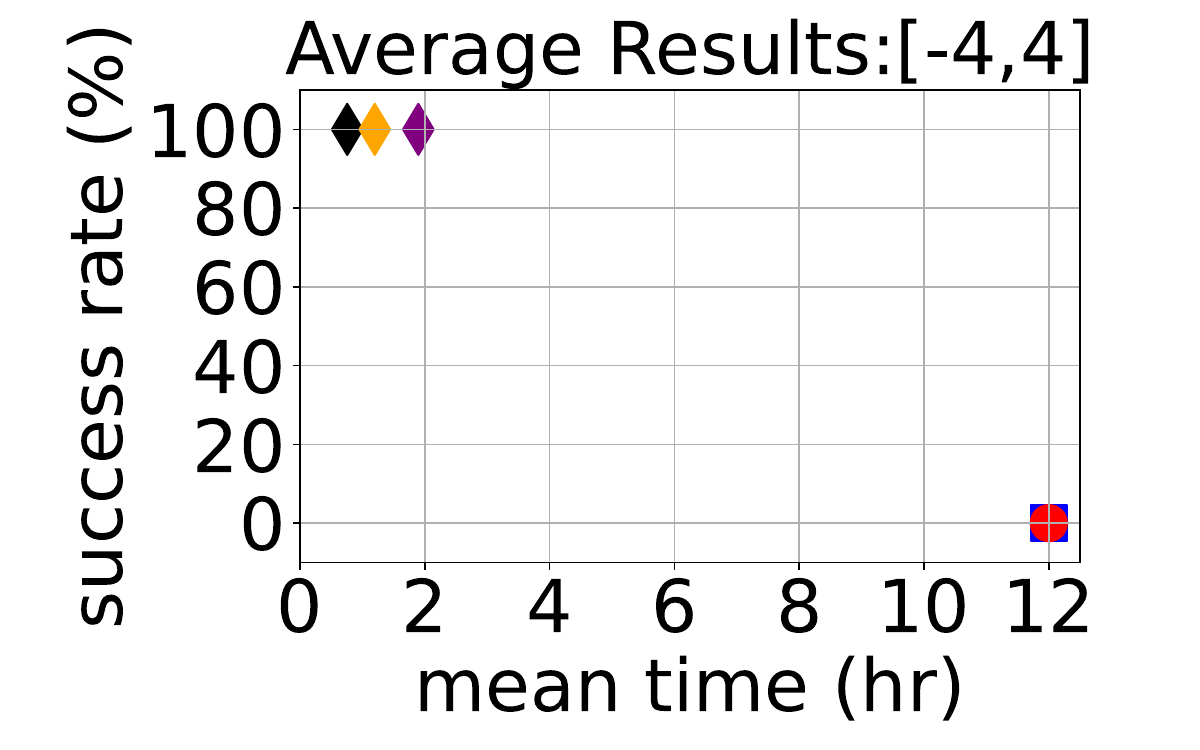}
    \label{fig:1d}
\end{subfigure}
\hfill
\begin{subfigure}{0.15\textwidth}
    \includegraphics[width=\linewidth]{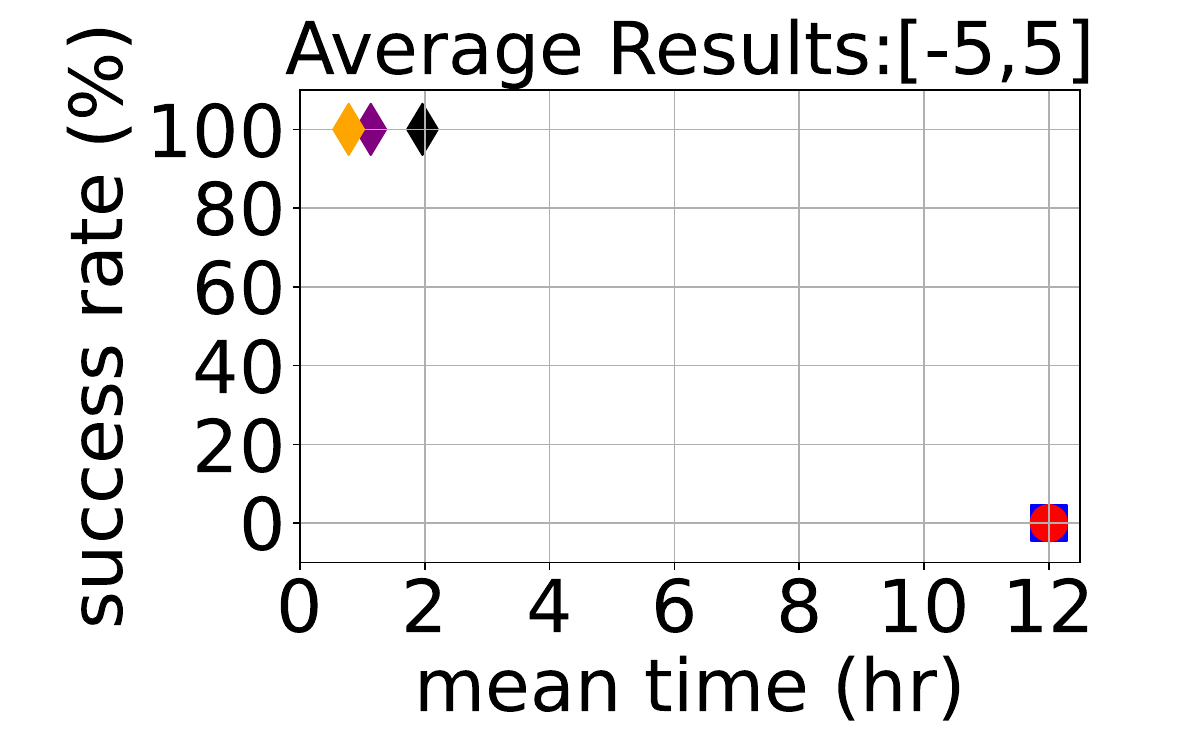}
    \label{fig:1e}
\end{subfigure}

\vspace{1em}  

\begin{subfigure}{0.15\textwidth} \includegraphics[width=\linewidth]{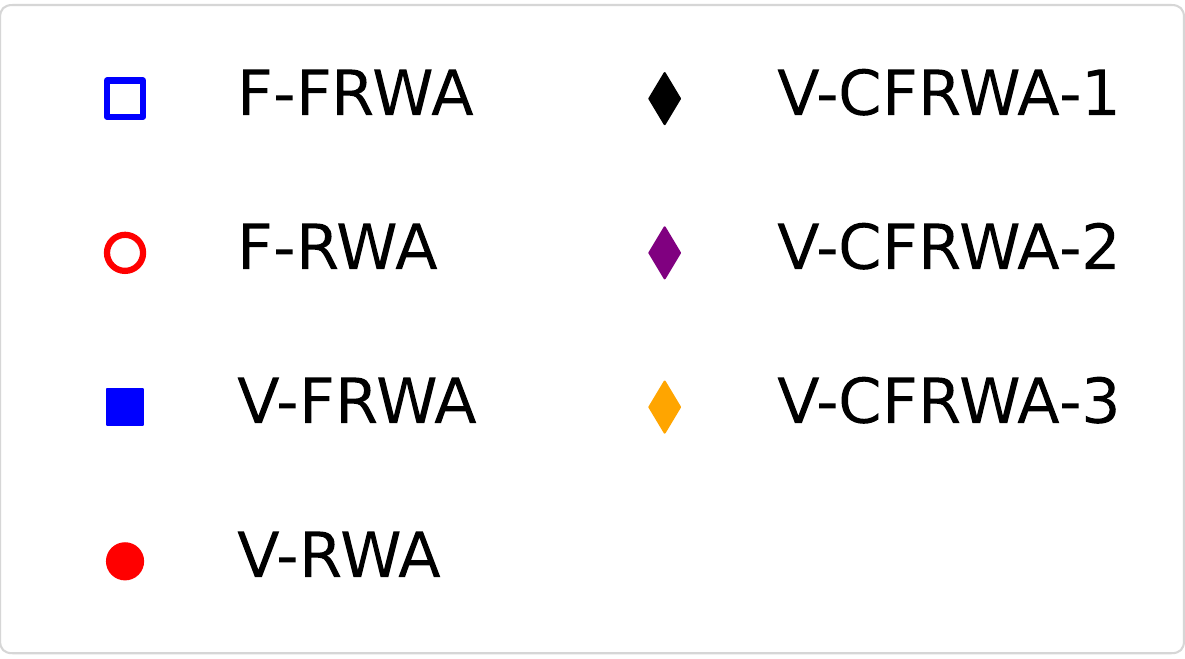} 
    \label{fig:label}
\end{subfigure}
\hfill
\begin{subfigure}{0.15\textwidth}
    \includegraphics[width=\linewidth]{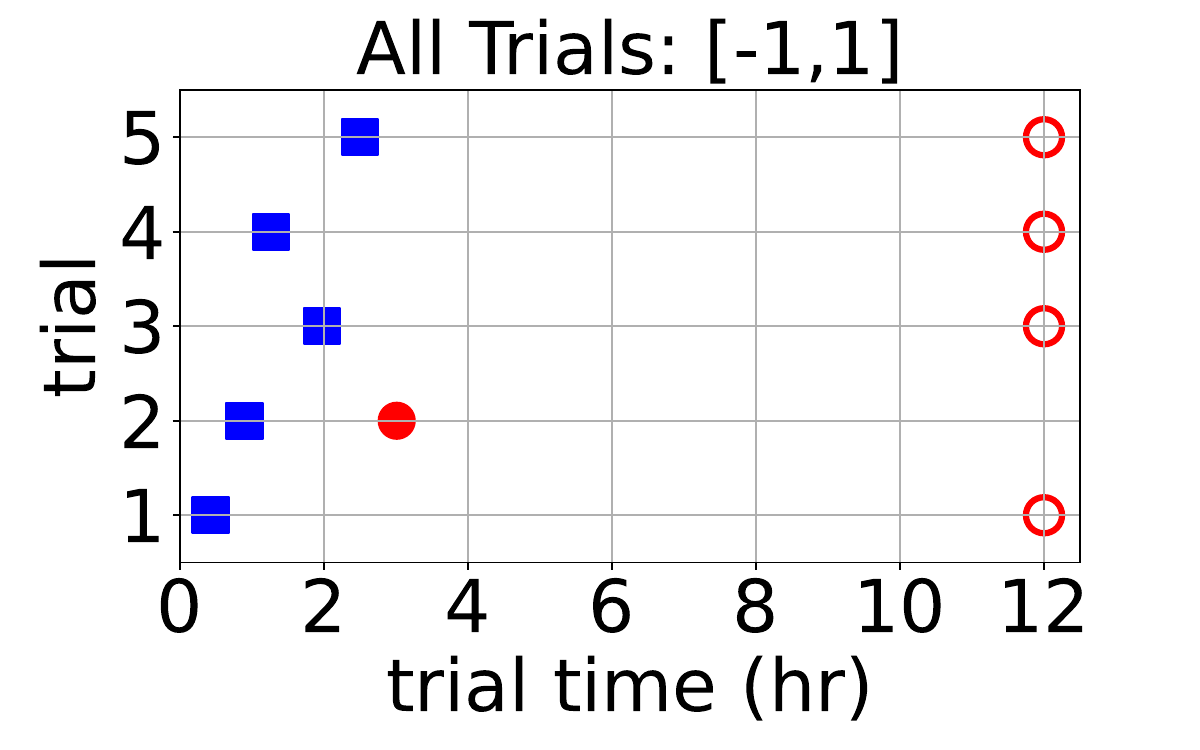}
    \label{fig:2a}
\end{subfigure}
\hfill
\begin{subfigure}{0.15\textwidth}
    \includegraphics[width=\linewidth]{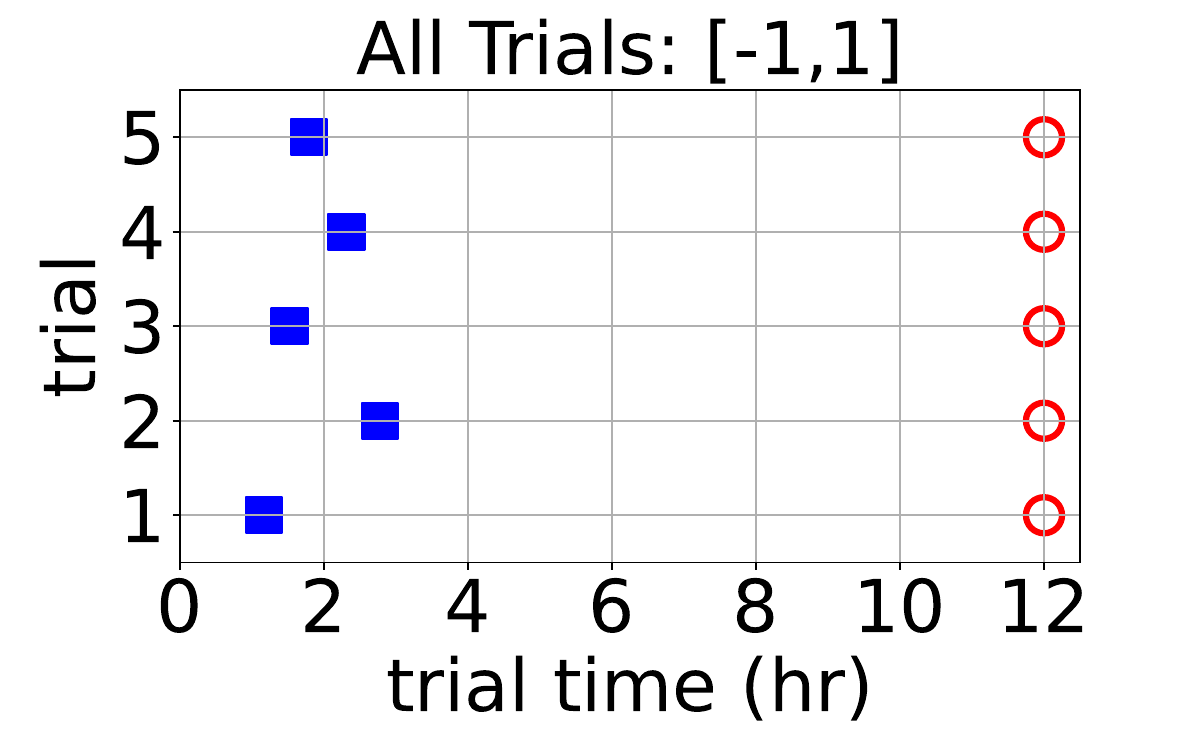}
    \label{fig:2b}
\end{subfigure}
\hfill
\begin{subfigure}{0.15\textwidth}
    \includegraphics[width=\linewidth]{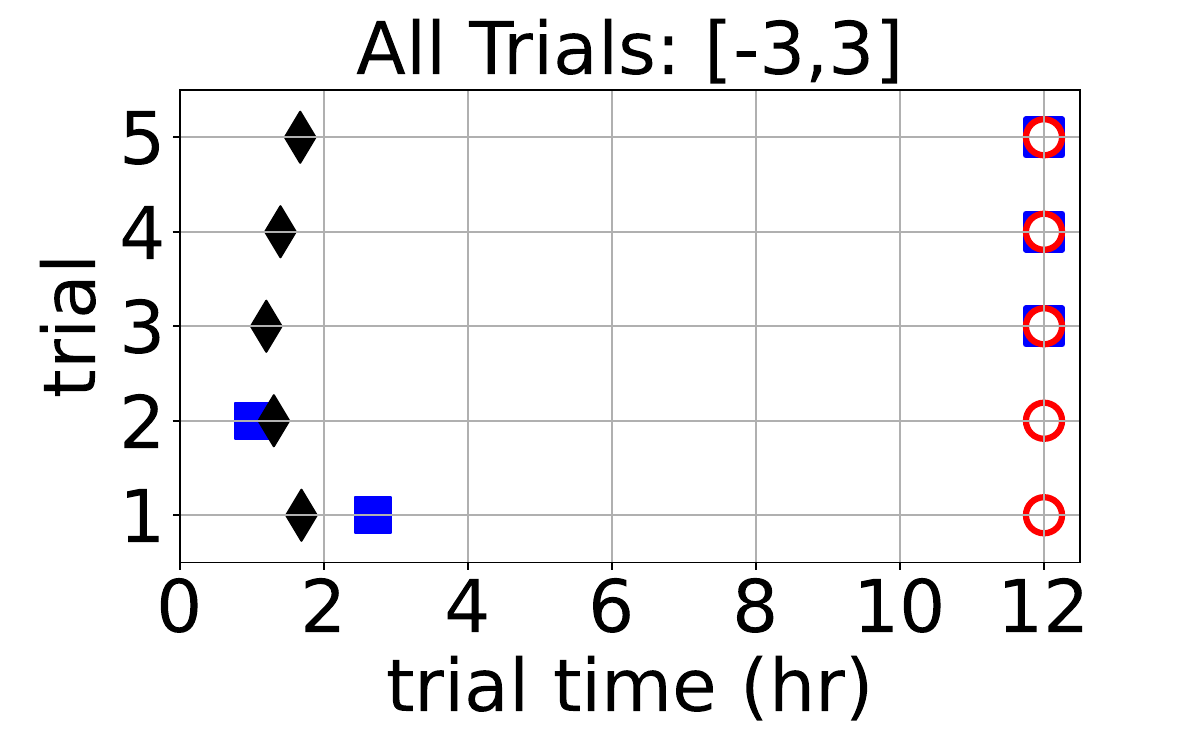}
    \label{fig:2c}
\end{subfigure}
\hfill
\begin{subfigure}{0.15\textwidth}
    \includegraphics[width=\linewidth]{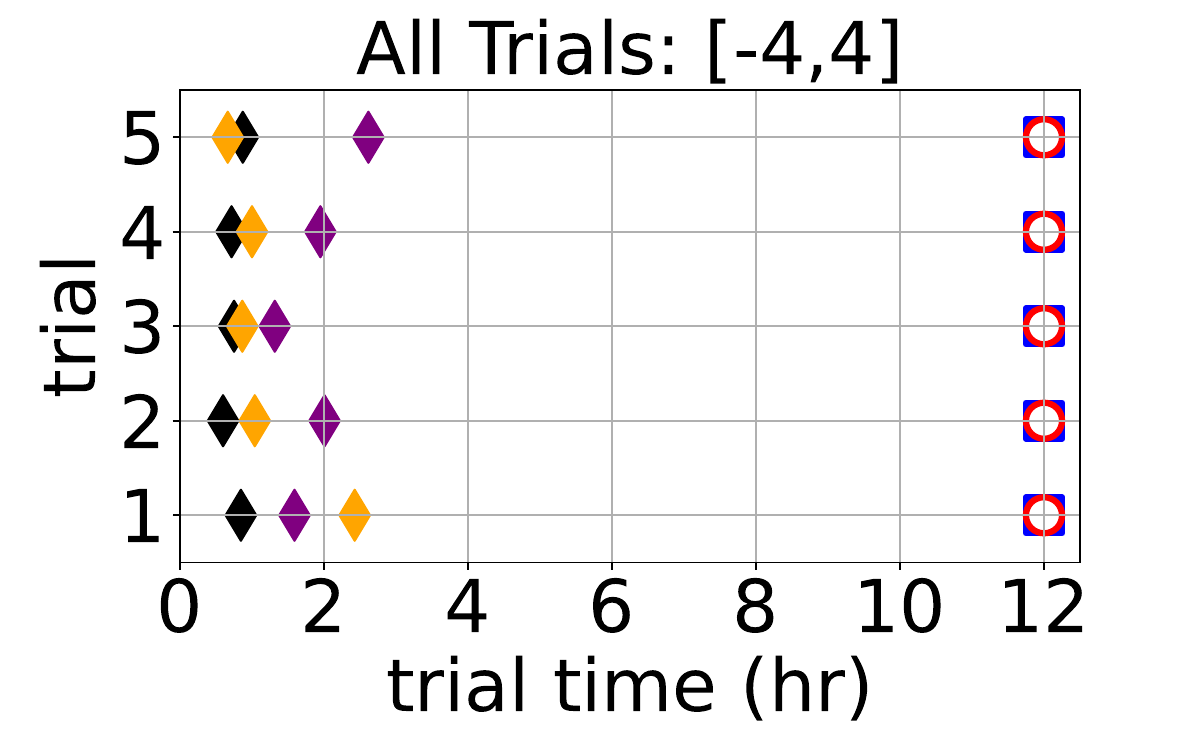}
    \label{fig:2d}
\end{subfigure}
\hfill
\begin{subfigure}{0.15\textwidth}
    \includegraphics[width=\linewidth]{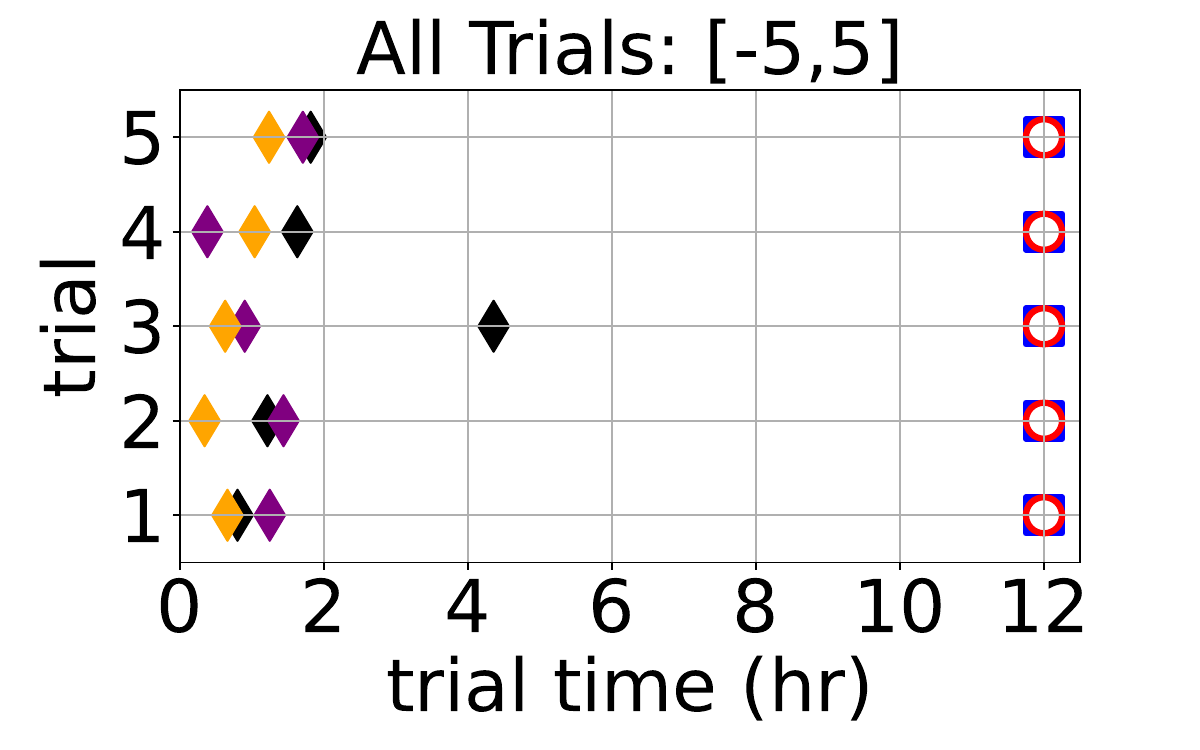}
    \label{fig:2e}
\end{subfigure}

\caption{The first 2 rows show average times and success rates for creating verified certificates over 5 trials. The bottom 2 rows show specific times for each trial, separated into failed (``F-'') and verified (``V-'') certificates. CFRWA-1, CFRWA-2, and CFRWA-3 refer to the first (up to) three CRWA tasks for the given starting region, corresponding, respectively to the first (up to) three rows for that starting region in Table~\ref{table:compositionalFRWA}.} \label{fig:composite}
\end{figure}

\paragraph{RWA vs. FRWA.}
In our first set of experiments, we select a set of RWA tasks and train both RWA and FRWA certificates using our CEGIS loop.\footnote{The Fossil 2.0 tool provides an implementation for computing the RWA certificates used in ~\cite{EdwPerAba2023}.  However, our definition of RWA is slightly different, and we use a different DNN verification tool, so we compare with our own implementation of RWA certificates to have a more meaningful comparison and to better isolate the contribution of the filtering technique.}
We select five RWA tasks, where $\states_I$ is set to $[-i,i]$ for the $i$th task.
For the RWA certificates, we follow the approach of~\cite{EdwPerAba2023}, whereas our FRWA certificates are constructed as described in Sec.~\ref{sec:approach}. In each case, we run five independent trials for each task.

The results are summarized in Fig.~\ref{fig:composite}.  The first two rows show, for each starting region, the number of successful runs (a run is successful if the CEGIS loop produces a fully verified controller/certificate pair within the 12 hour time limit) and the average time required for the successful runs.  Results for RWA are shown as red circles and FRWA as blue squares (we explain the diamonds later).  For example, for starting region $[-2,2]$, all five trials are successful for FWRA, with an average time of 2 hours, whereas all five trials are unsuccessful for RWA.  The bottom two rows show data from the same experiments, but here we show the time taken for each of the five trials.  An unfilled circle or box represents a timeout.

The results suggest that FWRA has a clear advantage over standard RWA.  In fact, RWA only succeeded once in producing any verified certificate, and only for the simplest starting region.  On the other hand, our FRWA approach is able to produce certificates faster and for starting regions up to $[-3,3]$.  After that, both techniques time out.

\begin{table*}[t]
\centering
\caption{Compositional certificate results. The columns indicate: the initial set for the final certificate, the size of the compositional certificate, the initial sets for all but the final certificate, the cumulative time for all but the final certificate, the total time (min, mean, and max) for training all certificates, and statistics for training the final certificate. We note that the cumulative time column is always equal to the corresponding value in the min column corresponding to the penultimate certificate. The wall time is the total time including the final controller/certificate. The CEGIS iterations and success stats are for the final controller/certificate only.}
\scalebox{1.0}{
\begin{tabular}{|c|ccc|ccc|ccc|c|}
\hline
&
 \multicolumn{3}{|c|}{Compositional Certificate}&
 \multicolumn{3}{|c|}{Wall Time (s)}&
  \multicolumn{3}{|c|}{CEGIS Iterations}
&
  \\
\hline
{$\mathbf{\states_I}$} & $\mathbf{n}$ & $\mathbf{\states^0_I\dots\states^{n-2}_I}$ & \textbf{Cumulative Time (s)} & \textbf{min(t)} & \textbf{mean(t)} & \textbf{max(t)} & \textbf{min(i)} & \textbf{mean(i)} & \textbf{max(i)} & \textbf{success (\%)} \\ \hline
[-2,2] & 1 & N/A & 0 & 4199 & 6890 & 8322 & 3 & 4.8 & 10 & 100 \\
\hline
[-3,3] & 1 & N/A & 0 & 3650 & 6644.5 & 9639 & 2 & 2.5 & 3 & 40 \\
\hline
[-3,3] & 2 & [-2,2] & 4199 & 8514 & 9421 & 10271 & 4 & 4.4 & 5 & 100 \\
\hline
[-4,4] & 2 & [-3,3] & 3650 & 5802 & 6374 & 6790 & 2 & 2.4 & 3 & 100 \\
\hline
[-4,4] & 2 & [-2,2] & 4199 & 8940 & 11026 & 13620 & 3 & 4.2 & 6 & 100 \\
\hline
[-4,4] & 3 & [-2,2], [-3,3] & 8514 & 10901 & 12829 & 17248 & 2 & 4.2 & 8 & 100 \\
\hline
[-5,5] & 2 & [-3,3] & 3650 & 6526 & 10716 & 19331 & 2 & 4.4 & 9 & 100 \\
\hline
[-5,5] & 3 & [-3,3], [-4,4] & 5802 & 7171 & 9884 & 11945 & 1 & 3.4 & 5 & 100 \\
\hline
[-5,5] & 4 & [-2,2],[-3,3],[-4,4] & 10901 & 12130 & 13710 & 15353 & 1 & 2.4 & 4 & 100 \\
\hline
[-6,6] & 3 & [-2,2],[-4,4] & 8940 & 13183 & 16384 & 20059 & 4 & 4.4 & 5 & 100 \\
\hline
[-6,6] & 4 & [-3,3],[-4,4],[-5,5] & 7171 & 9680 & 14027 & 32103 & 2 & 4.6 & 9 & 100 \\
\hline
[-6,6] & 5 & [-2,2],[-3,3],[-4,4],[-5,5] & 12130 & 18607 & 21768 & 24356 & 3 & 4.4 & 5 & 100 \\
\hline
[-7,7] & 3 & [-3,3],[-5,5] & 6526 & 9158 & 10171 & 10848 & 2 & 2.8 & 3 & 100 \\
\hline
[-7,7] & 5 & [-3,3],[-4,4],[-5,5],[-6,6] & 9680 & 11878 & 15967 & 23419 & 2 & 3.6 & 7 & 100 \\
\hline
[-8,8] & 4 & [-2,2],[-4,4],[-6,6] & 13183 & 16677 & 22623 & 33849 & 2 & 3.2 & 4 & 100 \\
\hline
[-9,9] & 4 & [-3,3],[-5,5],[-7,7] & 9158 & 12919 & 16013 & 18507 & 2 & 3.4 & 5 & 100 \\
\hline
[-10,10] & 5 & [-2,2],[-4,4],[-6,6],[-8,8] & 16677 & 18137 & 23421 & 30872 & 1 & 3.4 & 6 & 100 \\
\hline
[-11,11] & 5 & [-3,3],[-5,5],[-7,7],[-9,9] & 12919 & 20641 & 27860 & 32834 & 1 & 2.6 & 5 & 100 \\
\hline
\end{tabular}
}\label{table:compositionalFRWA}
\end{table*}

\paragraph{Compositional Certificates.}
As demonstrated above, RWA and FRWA certificates quickly run into scalability challenges on our case study problem.  For example, even with 5 tries and a 12 hour timeout, neither approach could produce a verified controller for the
$[-4,4]$ or $[-5,5]$ starting regions.

Our second set of experiments demonstrates that this scalability challenge can be addressed with compositional certificates.  We train a set of compositional certificates (each composed of multiple FRWA certificates) and report the results in Table~\ref{table:compositionalFRWA}.

Each row of the table corresponds to a compositional certificate.  The first column shows the value of $\states_I$ for this certificate.  The next columns indicate the number $n$ of composed certificates, the values of $\states^i_I$ for $0\le i < n-1$, and the cumulative time required for all but the last certificate.  The next three columns give the minimum, mean, and maximum time required to produce the controller and certificate for the last stage of the compositional certificate (recall that we run five independent trials for all CEGIS loops).  The next three columns show the minimum, mean, and maximum number of CEGIS iterations used, and the last column indicates how many of the trials succeeded. Note that when $n=1$, the row corresponds to a single FRWA certificate.

The results clearly indicate that compositional certificates greatly improve scalability.  Whereas the stand-alone certificates could not scale beyond $[-3,3]$ in 12 hours, we were able to successfully produce a formally verified $5$-stage certificate for $[-11,11]$ in a little over $5.7$ hours.  It is also worth noting that we do get a significant benefit by running 5 independent CEGIS loops, as both the time and the number of loops can vary significantly from the minimum to the maximum.  Nearly all of the CEGIS loops eventually completed---only the initial $[3,3]$ region failed to complete all of its trials---suggesting that the compositional approach is also more stable and robust.  This can also be seen in \cref{fig:composite}: for each starting region $[-a,a]$, the diamond point labeled CFRWA-$i$ corresponds to the $i$th row containing $[-a,a]$ in column 1.  We can see that, compared to the stand-alone RWA and FRWA certificates, the compositional certificates can be trained faster and with fewer failures.

\section{Conclusion}
\label{sec:Conclusion}

In this work, we present a novel framework for formally verifying DRL-based controllers. 
Our approach leverages Neural Lyapunov Barrier certificates and demonstrates how they can be used to verify DNN-based controllers for complex systems.
We use a CEGIS loop for training and formally verifying certificates, and we introduce filters for reach-while-avoid certificates, which simplify the training and verification process.
We also introduce compositional certificates which use a sequence of simpler certificates to scale to large state spaces.

We demonstrate the merits of our approach on a 2D case study involving a DRL-controlled spacecraft which is required to dock in a predefined region, from any initialization point. We demonstrate that for small subdomains, our FRWA approach is strictly better than competing RWA-based certificate methods. Furthermore, we demonstrate that our compositional approach unlocks significant additional scalability.

In the future, we plan to extend our approach to be compatible with additional formal techniques (e.g., shielding against safety violations~\cite{WuMaDeWa19, BlKoKoWa15, AlBloEh18,PraKoTa21,KoLoJaBl20,CoAmRoSaKaFo24,RoAmCoSaKa24}, and Scenario-Based Programming~\cite{GoMaMe12Spaghetti,katzElyasaf2021DLSBP,GrGrKaMa16,KaMaSaWe19,K20,CoYeAmFaHaKa22,YeAmElHaKaMa22,YeAmElHaKaMa23}). 
We also plan to apply our approach to more challenging case studies with larger DRL controllers.
We see this work as an important step towards the safe and reliable use of DRL in real-world systems.


\section{Acknowledgements}
\label{sec:Acknowledgements}
This work was supported by AFOSR (FA9550-22-1-0227), the Stanford CURIS program, the NSF-BSF program (NSF: 1814369, BSF: 2017662), and the Stanford Center for AI Safety.
The work of Amir was further supported by a scholarship from the Clore Israel Foundation.
We thank Kerianne Hobbs (AFRL), Thomas Henzinger (ISTA), 
Chuchu Fan (MIT), and Songyuan Zhang (MIT) for useful conversations and advice which contributed to the success of this project.

\newpage
{
\bibliographystyle{abbrv}
\bibliography{references}
}

\newpage
\onecolumn
\setcounter{section}{0}

\clearpage
{\huge{Appendix}}

\section{Discrete Time Step State Computation For 2D Docking Spacecraft}\label{sec:appendix:discretecomp}

For the spacecraft, the next state $\boldsymbol{\state}_i' = [x', y', \dot{x}', \dot{y}']$ given by a discrete time-step of $T$ from the previous state $\boldsymbol{\state} = [x, y, \dot{x}, \dot{y}]$ and control inputs $\boldsymbol{u} = [F_x, F_y]$, with spacecraft mass $m = 12$ kg and constant $n = 0.001027$ rad/s is:

\begin{align}
x' = (\frac{2 \dot{y}}{n} + 4 x + \frac{F_x}{m n^2}) + (\frac{2 F_y}{mn})
+ (- \frac{F_x}{m n^2} - \frac{2 \dot{y}}{n} - 3 x)\cos{(nT)} +    (\frac{-2 F_y}{m n^2} + \frac{\dot{x}}{n})\sin{(nT)}\\
y' = (-\frac{2\dot{x}}{n} + y + \frac{4 F_y}{m n^2}) + (\frac{-2 F_x}{mn} - 3\dot{y} - 6nx)T + -\frac{3F_y}{2m}t^2 + (-\frac{4F_y}{mn^2} + \frac{2\dot{x}}{n})\cos{(nT)} + (\frac{2F_x}{mn^2} + \frac{4\dot{y}}{n} + 6x)\sin{(nT)}\\
\dot{x}' = (\frac{2F_x}{mn}) + (\frac{-2F_y}{mn} + x)\cos{(nT)} + (\frac{F_x}{mn} + 2\dot{y} + 3nx)\sin{(nT)}\\
\dot{y}' = (\frac{-2F_x}{mn} - 3\dot{y} - 6nx) + (-\frac{3 F_y}{m})T + (\frac{2F_x}{mn} + 4\dot{y} + 6nx)\cos{(nT)} + (\frac{4 F_y}{mn} - 2 \dot{x})\sin{(nT)}
\end{align}

\newpage

\section{DNN Architecture \& Training}
\label{sec:appendix:DnnTraining}


\subsection{Initial Training Phase}
The neural network controller is initialized by training on the 2D docking scenario using the popular \emph{Proximal Policy Optimization} (PPO) RL algorithm implemented by Ray RLlib in the same manner as the original SafeRL benchmark~\cite{ravaioli2022safe}. However, a number of changes were made to the network and training procedure to make the network more amenable to verification, including:

\subsubsection{Scenario}: The initial distance from the chief spacecraft was reduced from 150m to 5m since we focus on verification at shorter distances. The docking region radius was reduced from 0.5m to 0.25m to avoid encountering boundary conditions during episode termination, while attempting to verify a 0.5m goal region. 

\subsubsection{Architecture}: The original \emph{tanh} activation functions were replaced with \emph{ReLU} activation functions in order to make the network piecewise linear, and hence compatible with our verifier. 
In addition, the hidden layer width was reduced from 256 to 20 neurons, per each layer. 
We note that the original network depth, i.e., the use of two hidden layers, was retained.

\subsubsection{Observations}: The observations of both (i) the scalar speed, and (ii) the distance-dependent velocity limit, were removed. This is due to their non-linear characterization with regard to the system's state space. However, the x, y position and velocity features were retained in the agent's observation space. It is important to note that the distance-dependent velocity limit can be reintroduced and learned through a post-training process, like the RWA training defined in our work.

\subsubsection{Rewards}: All the rewards described in the original benchmark were retained including the distance change, delta-v, safety constraint, and success/failure rewards. However, we significantly modified the reward pertaining to the change in the spacecraft's distance. Specifically, the distance metric was changed from the previously-used Euclidean distance to a Manhattan distance based on the L1 norm of the position vector. This change resulted in behavior that was easier to verify with a linear verification constraint. The reward was also re-calibrated to account for the novel initial distance value (5m) and was augmented with an additional distance update reward term that was calibrated with a midpoint at 0.5m. This value was updated in order to encourage the agent's location to to convergence close to the goal as well as for avoiding spiral trajectories. The updated distance reward consisted of the following form:

\begin{align} 
    R^{d_{new}}_t &= 2 \left( e^{-a_1d^m_{t}} - e^{-a_1d^m_{t-1}} \right) + 2 \left( e^{-a_2d^m_{t}} - e^{-a_2d^m_{t-1}} \right) \\ 
    d^m_i &= \left\lvert x_i \right \rvert + \left\lvert x_i \right \rvert \\ 
    a_1&=\frac{\ln(2)}{5},\quad a_2=\frac{\ln(2)}{0.5} 
\end{align}

\begin{figure}[h]
    \centering
    \includegraphics[width=0.5\textwidth]{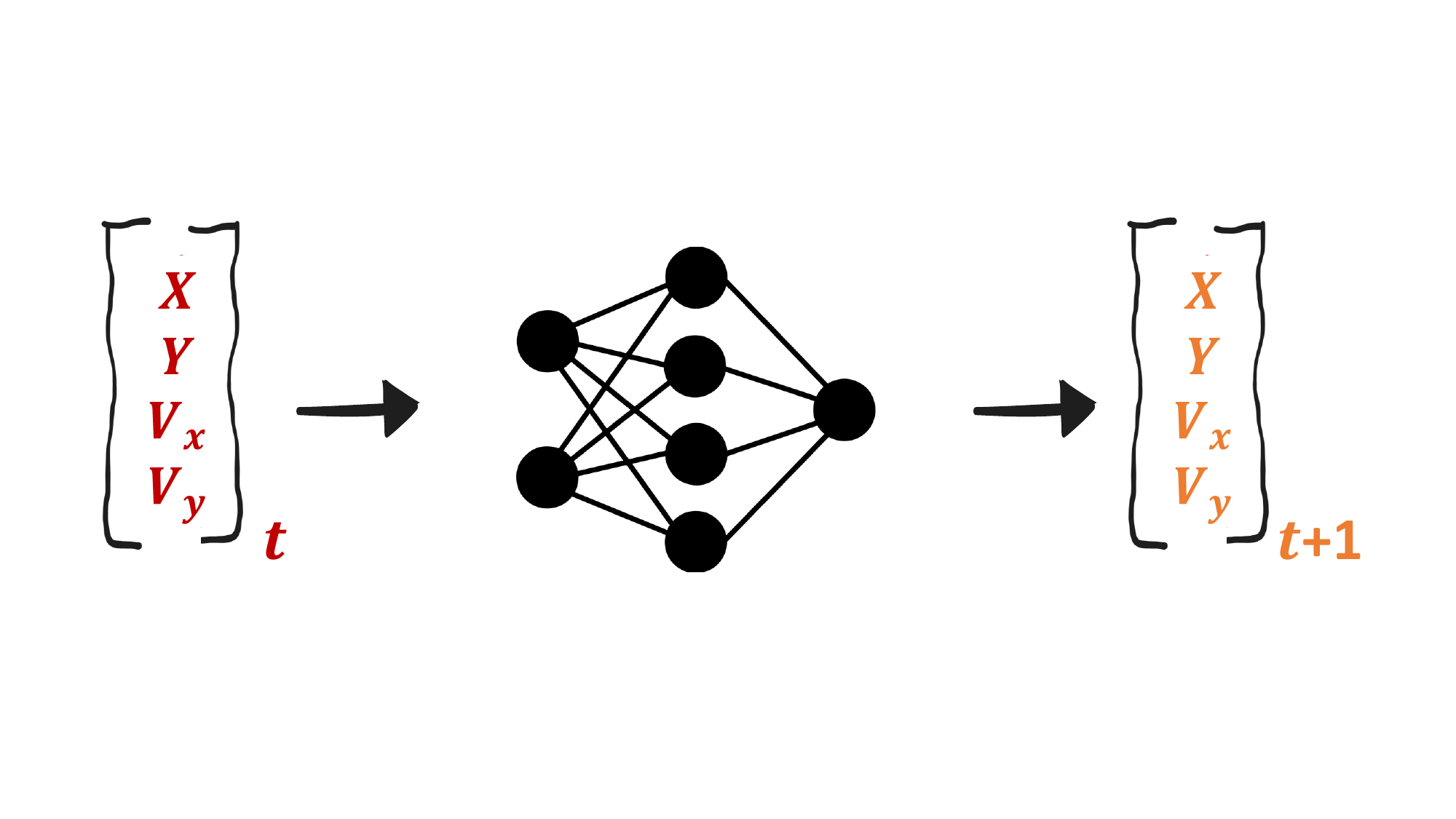}
    \caption{A scheme of the DNN controller architecture for the 2D docking benchmark. Given an input state of the system, $\boldsymbol{x} = [x_{t}, y_{t}, \dot{x_{t}}, \dot{y_{t}}]^T$, the DNN outputs the forces $[F_x,F_y]$, which are converted using dynamics $f$ to produce the next state. The states generated in this trajectory pertain to the deputy spacecraft, which attempts to safely maneuver into close proximity of a chief spacecraft.}
    \label{fig:docking_general_plot_new}
\end{figure}

\newpage

\section{Proofs}
\label{sec:appendix:proofs}

\setcounter{lemma}{0}
\begin{lemma}\label{lem:1}
{
If \certfun is an RWA certificate for a dynamical system with witness $(\alpha,\beta,\epsilon)$, and $V$ has a lower bound, then for every infinite trajectory $\traj$ starting from a state $\state\in\states\setminus\states_G$ such that $\certfun(\state)\le \beta$, $\traj$ will eventually contain a state in $\states_G$ without ever passing through a state in $\states_U$.} 
\end{lemma}
\textit{
{Proof.} Let $\psi$ be the minimum of $V(\state)$ for $\state\in\states$. Let us consider any initial state $x$ that meets the properties $V(\state) \le \beta, \state \in \states \setminus (\states_G)$. Note that $\states_I \setminus \states_G$ is a subset of the set of initial states. Therefore we additionally claim that all states starting from starting states which are not already in the goal region will also reach the goal region, by completing this proof. Returning to the proof, the next state given by the discrete time-step system, $\state'$, must either be in $\states_G$ or it must be the case that $V(\state') \le \beta$.
Hence, it is straightforward to show that the trajectory starting from a defined initial state controlled by the certified controller $\pi$ will either reach a state in $\states_G$ or that it will produce a state $\state'$ such that $V(\state') \le \beta - t \epsilon$ where $t$ denotes the number of time-steps in between the starting state of the trajectory and $\state'$. As such, $\state'$ will provably also reach a value less than $\psi$ if the trajectory continues for an excessive number of time-steps while not reaching a state in $\states_G$, which is a contradiction. Thus, the trajectory starting from $\states_I$ or from state $\state$ such that $V(\state) \le \beta, \state \in \states \setminus \states_G$ must reach a state in $\states_G$ in, at most, $(\beta - \psi)/\epsilon$ time-steps. Moreover, since along the trajectory it must be the case that all states have values from $V$ be strictly less than $\beta$, and the value of $V$ for any state in $\states_U$ is at least $\alpha$, where $\alpha > \beta$, then for no state in the trajectory is it possible for the state to be in $\states_U$.}

\begin{lemma}
{Given a CRWA certificate for an RWA task with parameters $\states_I$, $\states_G$, and $\states_U$, all trajectories guided by the meta-controller starting at any point in $\states_I$ will reach $\states_G$ in a finite number of steps while avoiding $\states_U$. In other words, a CRWA certificate provides a correct solution for the RWA task.}
\end{lemma}
\textit{{Proof.} From RWA guarantees, it is easy to see that a verified RWA or FRWA controller $\pi_i$ (for $0 \le i \le n-1$) guides a trajectory starting from $\state \in \states \setminus \states^i_G$ where $V(\state) \le \beta_i$, then a state in $\states^i_G$ is reached in a finite number of states, and further, no state in $\states^i_U$ is ever reached.\\
Now, we know that trajectories begin under the control of controller $\pi_{i}$, and start at a state in $\states^{i}_I$ where $i \le n-1$. Since states in $\states^{i}_I$ have values produced from $V_{i}$ be $\le \beta_{i}$, then we know a state in $\states^{i}_G$ must be reached, while all states in $\states^{i}_U$ are avoided. \\
Note that $\states_G^{i}$ consists of states $x$ which are either in $\states^0_G$ or are in $\states \setminus (\states^0_G \cup \states^{i-1}_U)$ with $V_{i-1}(x) \le \beta_{i-1}$. As a result of RWA guarantees, we can then certify that if $\states^0_G$ is not already reached, the next controller $
\pi_{i-1}$ begins either (1) operation at a state $x$ such that $x \in \states / \states^{i}_G$, at which the the controller can continue guiding the state $x$ to a state in either $\states^0_G$ or $\states^{i-1}_G$, avoiding unsafe states in $\states^{i-1}_U$ (2) or otherwise the controller begins at a state in $\states^{i-2}_G$. In either case, it is clear that if $\states^0_G$ is not already reached, the following controller $\pi_{i-2}$ can continue guiding the system safety until its respective goal region is reached, and this operation can continue until $\states^0_G$ is reached.\\
It is clear that for each controller, $\pi_i$, that there is safe control and that all states in $\states^i_U$ are avoided for the relevant controller. Moreover, we see that for all trajectories starting from a state in $\states^{n-1}_I$, that a state in $\states^0_G$ is eventually reached. Both of these properties clearly show RWA  guarantees over goal region $\states^0_G$, unsafe region $\states^{n-1}_U$, and starting region $\states^{n-1}_I$, since all $\states^i_U \supseteq \states^{n-1}_U$.}

\newpage
\section{Failed Verification Attempts}
\label{sec:appendix:failedVerificationAttepts}

\subsection{$k$-induction}
Typical verification with $k$-induction is challenging, since (1) properties which provide docking guarantees need be determined, and (2) these properties must be verified with a value of $k$ which is tractable for DNN verifiers like Marabou. Some attempts at initially verifying liveness properties for out spacecraft (specifically reaching the docking region in a trajectory) was done by finding properties which implied docking guarantees which ideally held over small values of $k$. However, we found a behavior for verifying these properties over $k$ steps for $k$-induction which is shown by the following figure:

\begin{figure}[h]
    \centering
    \includegraphics[width=0.4\textwidth]{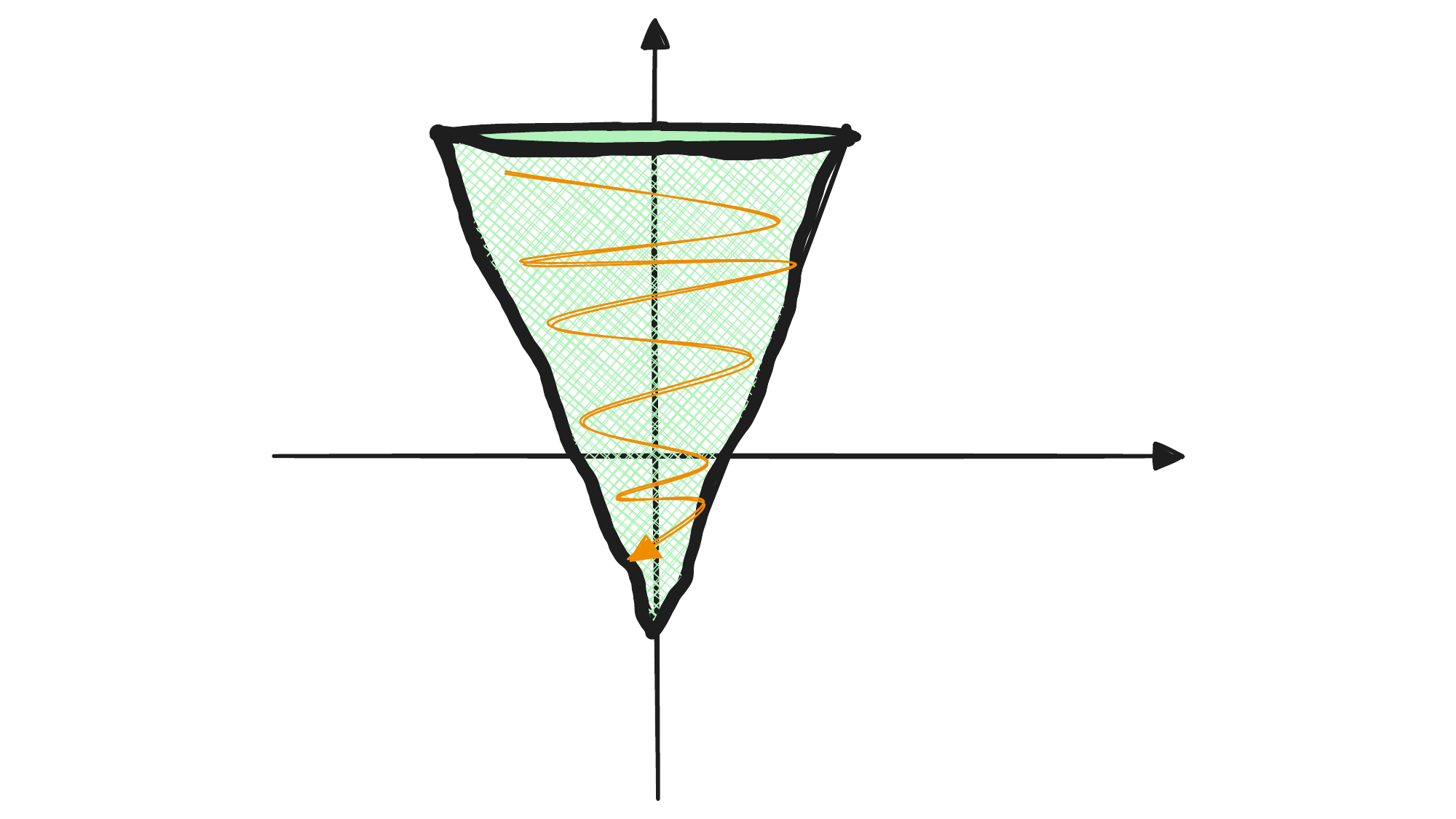}
    \caption{Behavior of $k$-induction with found properties. The conical region denotes where counterexamples typically occur for properties (e.g. decreasing distance). Typically, as the spacecraft is closer to the origin on one axis for position, the behavior does not hold as the spacecraft must make elaborate manuevers to guide towards the docking region, which is near the origin. This common occurence of counterexamples in the conical region limits the practicality of using $k$-induction.}\label{fig:coneFailedAttempt}
\end{figure}

\subsection{Grid reachability}
An alternate approach to verifying liveness properties for our spacecraft is by taking an abstraction-based approach. In this approach, the state space over which the spacecraft can operate over is split into grid cells, over which edges are determined between cells. A directed edge from cell $A$ to cell $B$ exists if there exists a point in cell $A$ can transition to cell $B$, for all cells outside of the docking region. Then after drawing up all grid cells and associated edges, it can be determined if there exist no cycles and edges leading out of the defined state space for edges from states inside the starting region, then liveness properties hold. However, in practice, this abstraction leads to cycles even when liveness holds, and even when grid cells are sufficiently small. This is because the behavior of states with velocities near 0 is somewhat unpredictable, especially for states near the docking region.

\begin{figure}[h!]
    \centering
    \includegraphics[width=0.4\textwidth]{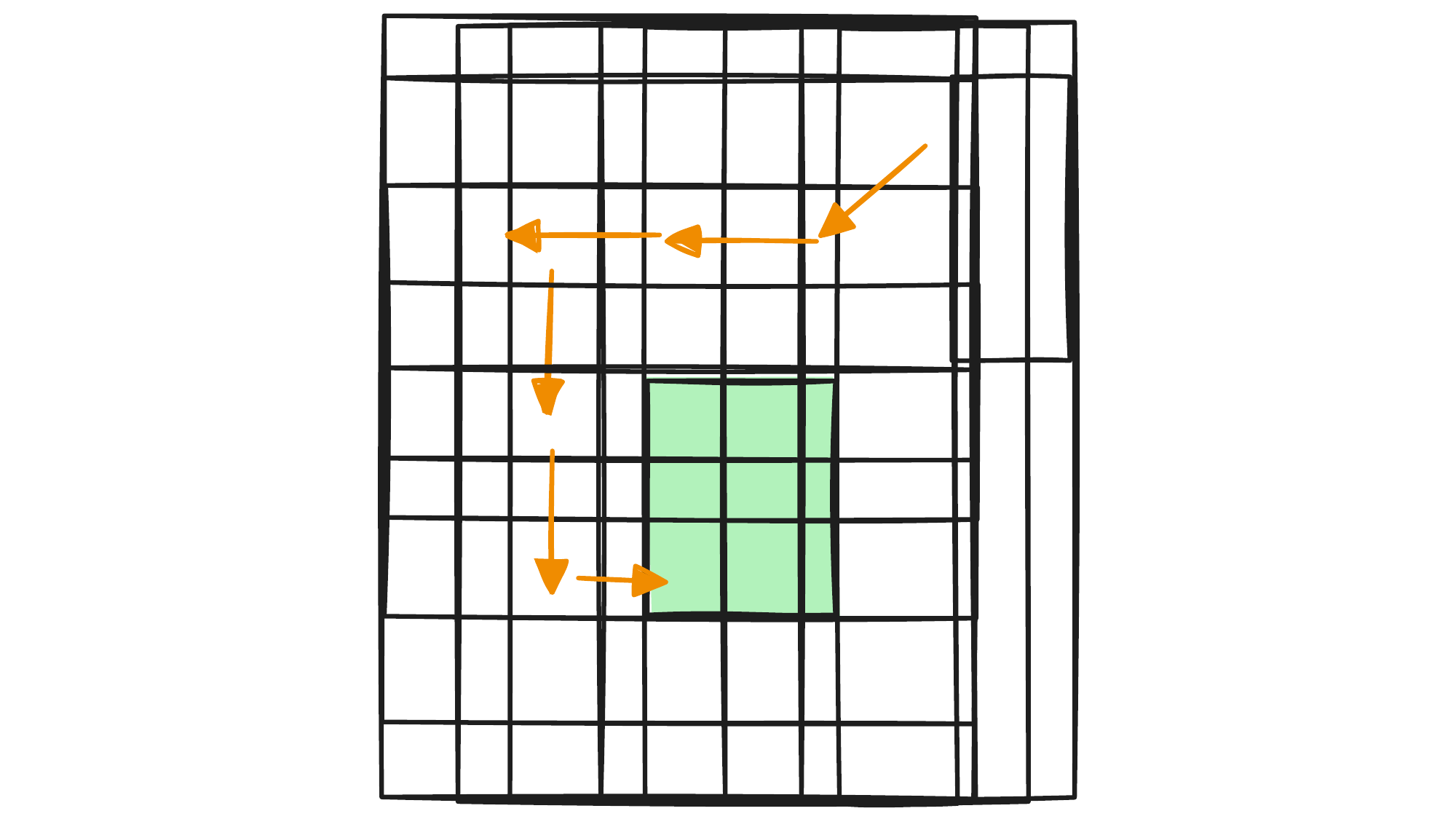}
    \caption{A possible trajectory found using verification over the abstracted grid cell space, where the docking region (green) is reached.}
    \label{fig:gridFailedAttempt}
\end{figure}

\newpage
\section{The structure of re-designed neural network used for the spacecraft controller.}
\label{sec:appendix:k_ind_nn}

\begin{figure}[h]
    \centering
    \includegraphics[width=0.2\textwidth]{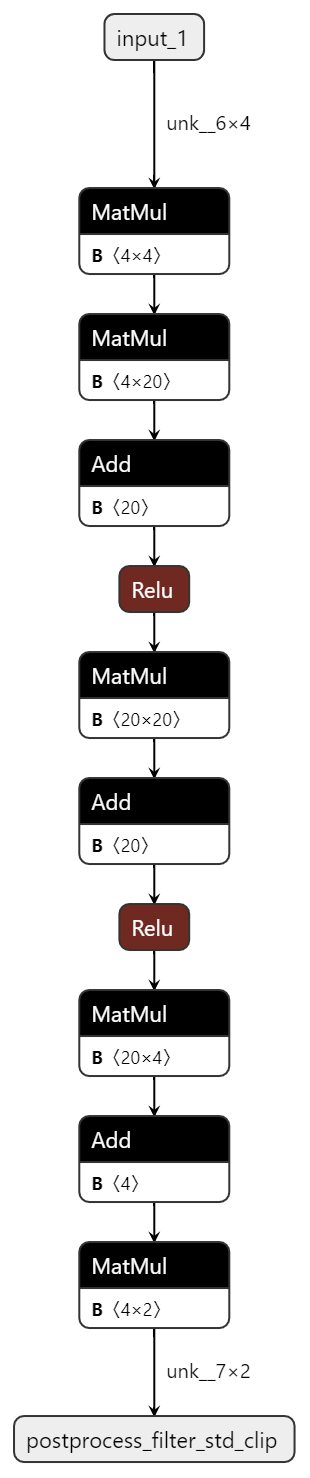}
    \caption{Visualized architecture of the redesigned neural network controller.\newline}
    \label{sec:appendix:onnx}
\end{figure}

The visualization of the retrained deep learning model reflects the structure and the data flow of the model, its layers and how they are connected. Each box represents a layer in the neural network. The upper and lower parts of the box indicate the layer type and the number of units (or filters) accordingly. \textbf{Input and output nodes}. Input box represents the shape of the input data (for example, $4\times4$ input is the 4 dimensional state. Output box represents the shape of the output data. For example, dimensions $unk\_7\times2$ are batch size of 7 and the output size of 2 (which is thrust on $x$ and $y$, components of the force). \textbf{Arrows} represent the connections between the layers and the direction of data flow.

\newpage
\section{Concrete Properties}
Here we formally present properties for which RWA guarantees should hold for the spacecraft. 
\subsubsection{Safe region} The safe region $x_s$ of the spacecraft is designed to consist of states where $\dot{x}, \dot{y} = 0$, and $-p < x,y < p$ for some constant position $p$. This is because the spacecraft always begins its trajectories with a velocity of 0.

\subsubsection{Goal region}
In order to give stronger and linear guarantees about reaching the docking region, the goal region $x_g$ is constrained by $-0.35 < x,y < 0.35$. Additionally, $\dot{x}$ and $\dot{y}$ are constraints by the safe velocity constraints discussed for $x_g$.

\subsubsection{Unsafe region}
The unsafe region $x_u$ is defined by states having unsafe velocities $\dot{x}$ and $\dot{y}$ in violations with the safe velocity constraints, and additionally consists of regions which are too far positionally for $x,y$ from the docking region. Specifically, either $|x|,|y| > u$ for some constant $u$.

\section{Linear Approximations}\label{sec:appendix:approx}
\subsection{Approximating the Euclidean Norm}

To verify the safety constraint presented in Equation \ref{eq:safe_constraint}, we must form a piecewise linear overapproximation of the constraint that is compatible with Marabou. Since the only non-linear components of the equations are Euclidean norms, we adopt the piecewise linear approximations developed in \cite{camino2019linearization}, where
\begin{align}
    \max_{i\in [1,n_{directions}]} (u_1 \cdot \cos(\frac{2(i-1)\pi}{n_{directions}}) + u_2 \cdot \sin(\frac{2(i-1)\pi}{n_{directions}})) \leq \sqrt{u_1^2+u_2^2}
\end{align}
and 
\begin{align}
    \frac{1}{cos(\pi/n_{directions})} \max_{i\in [1,n_{directions}]} (u_1 \cdot \cos(\frac{2(i-1)\pi}{n_{directions}}) + u_2 \cdot \sin(\frac{2(i-1)\pi}{n_{directions}})) \geq \sqrt{u_1^2+u_2^2}
\end{align}
for a positive integer $n_{directions}$. Note that we are able to pre-compute the values of $\cos(\frac{2(i-1)\pi}{n_{directions}})$ and $\sin(\frac{2(i-1)\pi}{n_{directions}})$ for $i\in [1, n_{directions}]$ before being passed to the SMT solver, thereby making the approximation linear. \\
For efficient computation, we restrict $n_{directions}$ to be a positive multiple of $4$ and utilize that \[ \sqrt{u_1^2+u_2^2} = \sqrt{|u_1|^2+|u_2|^2}\] and that
\[ |u_1| \cdot \cos(\frac{2(i-1)\pi}{n_{directions}}) + |u_2| \cdot \sin(\frac{2(i-1)\pi}{n_{directions}})\] 
is maximized on the interval of $i \in [1, n_{directions}/4+1]$, allowing us to equivalently rewrite that
\begin{align}
\label{eq:underapprox}
\text{under}(u_1, u_2)=
    \max_{i\in [1,n_{directions}/4+1]} (|u_1| \cdot \cos(\frac{2(i-1)\pi}{n_{directions}}) + |u_2| \cdot \sin(\frac{2(i-1)\pi}{n_{directions}})) \leq \sqrt{u_1^2+u_2^2}
\end{align}
and 
\begin{align}
\label{eq:overapprox}
    \text{over}(u_1, u_2)=
    \frac{1}{cos(\pi/n_{directions})} \max_{i\in [1,n_{directions}/4+1]} (|u_1| \cdot \cos(\frac{2(i-1)\pi}{n_{directions}}) + |u_2| \cdot \sin(\frac{2(i-1)\pi}{n_{directions}})) \geq \sqrt{u_1^2+u_2^2}.
\end{align}
By using only a fourth of the original intervals of $i$, we are able to significantly reduce the workload of the SMT solver. \\

\subsection{Overapproximating the Unsafe Region}
By the safety constraint defined in Equation \ref{eq:safe_constraint}, a state described by $\boldsymbol{\state}_t=[x_t, y_t, \dot{x}_t, \dot{y}_t]$ is in the unsafe region if 
\begin{align}
    \sqrt{\dot{x}_t^2+\dot{y}_t^2} > 0.2 + 2n \sqrt{x_t^2+y_t^2}.
\end{align}
We form a piecewise linear overapproximation of the unsafe region using Equations \ref{eq:underapprox} and \ref{eq:overapprox} such that state $\boldsymbol{\state}_t$ is unsafe if
\begin{align}
\label{eq:over_unsafe}
    \text{over}(\dot{x}_t, \dot{y}_t) \geq 0.2 + 2n \cdot \text{under}(x_t, y_t).
\end{align}
This overapproximation guarantees that a state in the unsafe space will always satisfy Equation \ref{eq:over_unsafe}.

\subsection{Direct Verification of the Safety Property}
\paragraph{}\label{paragraph:safety_linearized}
To directly verify the safety property presented in Equation \ref{eq:safe_constraint}, we note that the agent's initial state is guaranteed to satisfy the constraint and that the agent should satisfy the safety constraint at the next time step if it satisfies the safety constraint at the current time step. That is,
\begin{align}
\sqrt{\dot{x}_{t}^2+\dot{y}_{t}^2} \leq 0.2 + 2n \sqrt{x_{t}^2+y_{t}^2} \rightarrow \sqrt{\dot{x}_{t+1}^2+\dot{y}_{t+1}^2} \leq 0.2 + 2n \sqrt{x_{t+1}^2+y_{t+1}^2}
\end{align}
should always hold on the desired region. This means that a counterexample violating this property will satisfy
\begin{align}
\sqrt{\dot{x}_{t}^2+\dot{y}_{t}^2} \leq 0.2 + 2n \sqrt{x_{t}^2+y_{t}^2} \land \sqrt{\dot{x}_{t+1}^2+\dot{y}_{t+1}^2} > 0.2 + 2n \sqrt{x_{t+1}^2+y_{t+1}^2}.
\end{align}
We use Marabou to search this violation for satisfiability by forming a piecewise linear overapproximation with the approximations detailed in Equations \ref{eq:underapprox} and \ref{eq:overapprox}, 
\begin{align}
\label{eq:direct_verify}
    \text{under}(\dot{x}_t, \dot{y}_t) \leq 0.2 + 2n\cdot \text{over}(x_t, y_t) \land \text{over}(\dot{x}_{t+1}, \dot{y}_{t+1}) \geq 0.2 + 2n \cdot \text{under}(x_{t+1}, y_{t+1}).
\end{align}
The use of an overapproximation guarantees that Equation $\ref{eq:direct_verify}$ will be unsatisisfiable only if the controller always adheres to the safety constraint.

\section{Additional Figures}
These additional figures all concern the same spacecraft problem discussed in our experiments.
\subsection{Comparison of FRWA and RWA}
\begin{table*}[h!]
\centering
\caption{RWA and FRWA results. min(t), mean(t), max(t) refer to minimum, mean, and maximum computation of wall time. min(i), mean(i), max(i) refer to minimum, mean, and maximum computation of CEGIS iterations. These min, mean, max computations are only done over computation of successful certificates. 
S rate is the success percentage of finding certificates with no counterexamples among 5 trials, (N/A) means there are no such successful certificates.}

\scalebox{1}{
\begin{tabular}{|cc|ccc|ccc|c|}
\hline
& &
 \multicolumn{3}{|c|}{Wall Time}&
  \multicolumn{3}{|c|}{CEGIS Iterations}
&
  \\
\hline
\textbf{Starting} & \textbf{Certificate} & \textbf{min(t)} & \textbf{mean(t)} & \textbf{max(t)} & \textbf{min(i)} & \textbf{mean(i)} & \textbf{max(i)} & \textbf{S rate} \\
\hline
$[-1,1]$ & RWA & 10835 & 10835 & 10835  & 8 & 8 & 8 & 20 \\
\hline
$[-2,2]$ & RWA & N/A & N/A & N/A & N/A & N/A & N/A & 0 \\
\hline
$[-3,3]$ & RWA & N/A & N/A & N/A & N/A & N/A & N/A & 0 \\
\hline
$[-4,4]$ & RWA & N/A & N/A & N/A & N/A & N/A & N/A & 0 \\
\hline
$[-5,5]$ & RWA & N/A & N/A & N/A & N/A & N/A & N/A & 0 \\
\specialrule{2pt}{0pt}{0pt}
$[-1,1]$ & FRWA & 1529 & 5077 & 8995  & 2 & 5.6 & 10 & 100 \\ 
\hline
$[-2,2]$ & FRWA & 4199 & 6890 & 8322 & 3 & 4.8 & 10 & 100 \\
\hline
$[-3,3]$ & FRWA & 3650 & 6644.5 & 9639 & 2 & 2.5 & 3 & 40 \\
\hline
$[-4,4]$ & FRWA & N/A & N/A & N/A & N/A & N/A & N/A & 0 \\
\hline
$[-5,5]$ & FRWA & N/A & N/A & N/A & N/A & N/A & N/A & 0 \\
\hline
\end{tabular}
}\label{table:RWAvsFRWAextra}
\end{table*}

\subsection{CFRWA Performance Over Multiple FRWAs}
\begin{table}[h!]
\centering
\caption{Demonstration of Training of 10 CFRWAs, where every CFRWA is composed of an initial certificate over starting space [-2,2], and provides guarantees for starting space [-4,4]. Every 2 CFRWAs share the same initial FRWA, and the data for the initial FRWA is given in the last 3 columns. Data for the last FRWA in the CFRWA is given in second, third, and fourth columns. Iter. is an abbreviation for iterations.}
\scalebox{1.0}{

\begin{tabular}{|c|c|c|c|c|c|c|}
\hline
Trial & Cur. FRWA Wall Time & Cur. FRWA CPU Time & Cur. FRWA CEGIS Iter. & Init. FRWA Wall Time & Init. FRWA CPU Time & Init. CEGIS Iter. \\ \hline
1 	& 8113  	& 77354	& 4      	& 4199         	& 40083       	& 3            	\\ \hline
2 	& 8762  	& 82097	& 5      	& 4199         	& 40083       	& 3            	\\ \hline
3 	& 6855  	& 57427	& 4      	& 10019        	& 80468       	& 10           	\\ \hline
4 	& 9995  	& 96572	& 4      	& 10019        	& 80468       	& 10           	\\ \hline
5 	& 12633 	& 121146   & 6      	& 5477         	& 50175       	& 4            	\\ \hline
6 	& 19824 	& 210271   & 7      	& 5477         	& 50175       	& 4            	\\ \hline
7 	& 6048  	& 50009	& 4      	& 8322         	& 91451       	& 3            	\\ \hline
8 	& 10258 	& 128412   & 3      	& 8322         	& 91451       	& 3            	\\ \hline
9 	& 5776  	& 52497	& 3      	& 6433         	& 65251       	& 4            	\\ \hline
10	& 20716 	& 207350   & 8      	& 6433         	& 65251       	& 4            	\\ \hline
\end{tabular}
}
\label{tab:appendix-2}
\end{table}

\subsection{Minimum Time Comparison In Proving Larger Certificates with CFRWAs}

\begin{figure}[h!]
    \centering
\includegraphics[width=0.6\textwidth]{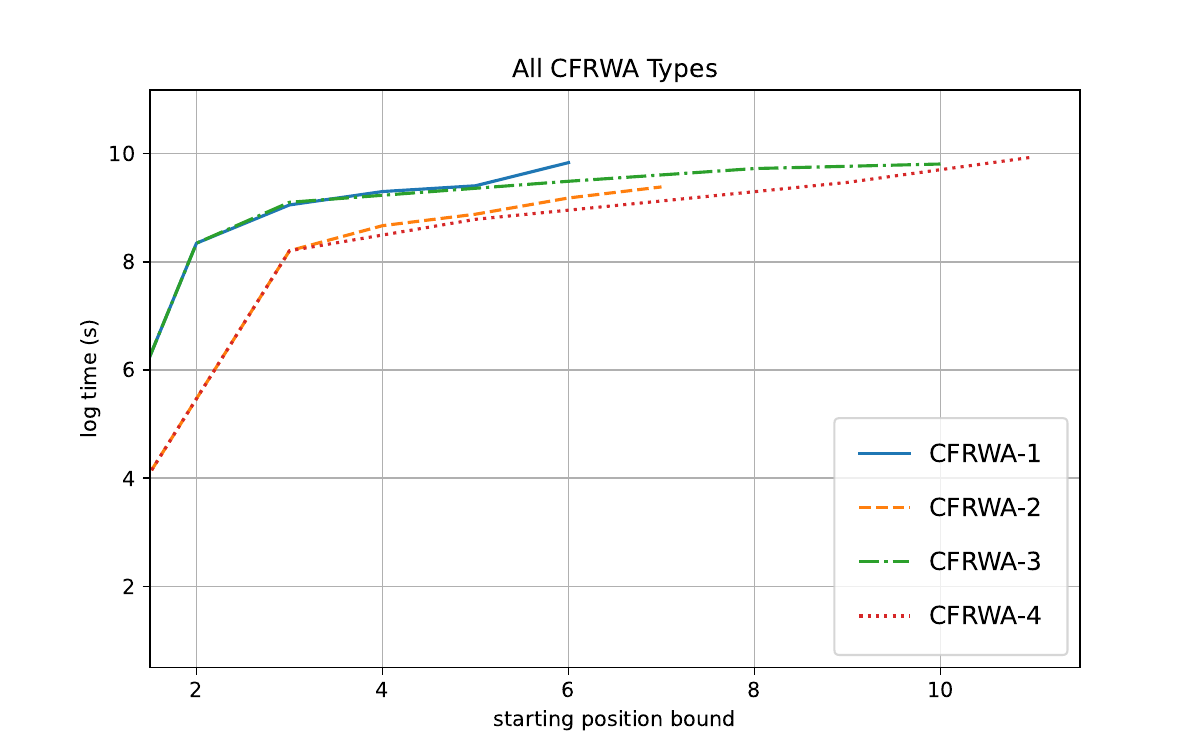} 
    \caption{Timing compared to position proven for four different configurations of CFRWA certificates. Each CFRWA certificate is composed of 5 FRWA certificates, each of which is learned iteratively to expand the certificate's state space of operation. The x-axis showcases the positional bound $[-d,d]$ with $d$ for the starting region that is proven by the iteration of CFRWA. CFRWA-1 begins with a FRWA with bound 2, and expands 1 meter with each successive FRWA. CFRWA-2 begins with a FRWA with bound 3, and expands 1 meter with each successive FRWA. CFRWA-3 begins with a FRWA with bound 2, and expands 2 meters at a time with each successive FRWA. CFRWA-4 begins with a FRWA with bound 3, and expands 2 meters at a time with each successive FRWA.}
    \label{fig:app-3}
\end{figure}

\end{document}